\documentclass[sigconf]{acmart}

\AtBeginDocument{%
  }

\newcommand{\SystemName}{UGAD}
\usepackage{supertabular, soul}
\usepackage{tabularx}
\usepackage{multirow}

\usepackage{microtype}
\usepackage{kantlipsum}
\usepackage{balance}  

\usepackage{graphicx}
\usepackage{lipsum}  
\usepackage[utf8]{inputenc}
\usepackage[T1]{fontenc}

\begin{document}

\title{UGAD: Universal Generative AI Detector utilizing Frequency Fingerprints}

\author{Inzamamul Alam}
\affiliation{%
  \institution{Sungkyunkwan University}
  \city{Suwon}
  \country{South Korea}
}
\email{inzi15@g.skku.edu}

\author{Muhammad Shahid Muneer}
\authornote{Equal Contribution.}
\affiliation{%
  \institution{Sungkyunkwan University}
  \city{Suwon}
  \country{South Korea}
}
\email{shahidmuneer@g.skku.edu}

\author{Simon S. Woo}
\authornote{Corresponding author.}
\affiliation{%
  \institution{Sungkyunkwan University}
  \city{Suwon}
  \country{South Korea}
}

\email{swoo@g.skku.edu}
\renewcommand{\shortauthors}{Inzamamul Alam, Muhammad Shahid Muneer, and Simon S. Woo}

\begin{abstract}

In the wake of a fabricated explosion image at the Pentagon, an ability to discern real images from fake counterparts has never been more critical. Our study introduces a novel multi-modal approach to detect AI-generated images amidst the proliferation of new-generation methods such as Diffusion models. Our method,~\SystemName, encompasses three key detection steps: First, we transform the RGB images into YCbCr channels and apply an Integral Radial Operation to emphasize salient radial features. Secondly, the Spatial Fourier Extraction operation is used for a spatial shift, utilizing a pre-trained deep learning network for optimal feature extraction. Finally, the deep neural network classification stage processes the data through dense layers using softmax for classification. Our approach significantly enhances the accuracy of differentiating between real and AI-generated images, as evidenced by a 12.64\% increase in accuracy and 28.43\% increase in AUC compared to existing state-of-the-art methods. 
  
\end{abstract}







\maketitle

\section{Introduction}

Rapid advancements in Generative AI have significantly impacted the digital realm, especially with the proliferation of content generation tools. This evolution has made creating and disseminating fake images easier, posing a challenge in distinguishing them from real ones and heightening the risk of misinformation. The advent of generative adversarial networks (GANs) and Stable Diffusion (SD) has notably enhanced AI's capability in photo-realistic content generation, presenting both creative opportunities and challenges, particularly in assuring content authenticity. As AI-generated content increasingly mimics reality, serious concerns have been raised over its potential misuse in activities. To address this, our paper focuses on detecting AI-generated fake images within the evolving AI landscape. 
Various deep-learning approaches, including pre-processing strategies to obtain power spectra for classification ~\cite{wu2023generalizable,corvi2023intriguing}, have been proposed to tackle this challenge. However, they fall short due to advancements in AI generation methods. Additionally, different forensic methods have been explored \cite{Corvi2022_on}. However, the accuracy and architectural adequacy of many existing fake image authentication methods need to be improved to effectively detect the latest AI-generated images. Benchmark datasets such as FaceForensics++ \cite{9010912}, CelebDF \cite{9156368}, and FakeAVCeleb~\cite{Khalid2021FakeAVCelebAN} have been instrumental in classifying images from varied generation methods. However, they fall short because they do not include the latest AI-generated datasets and may not be able to cope with those methods.

\begin{figure*}[ht]
  \centering
  \includegraphics[width=0.8\textwidth]{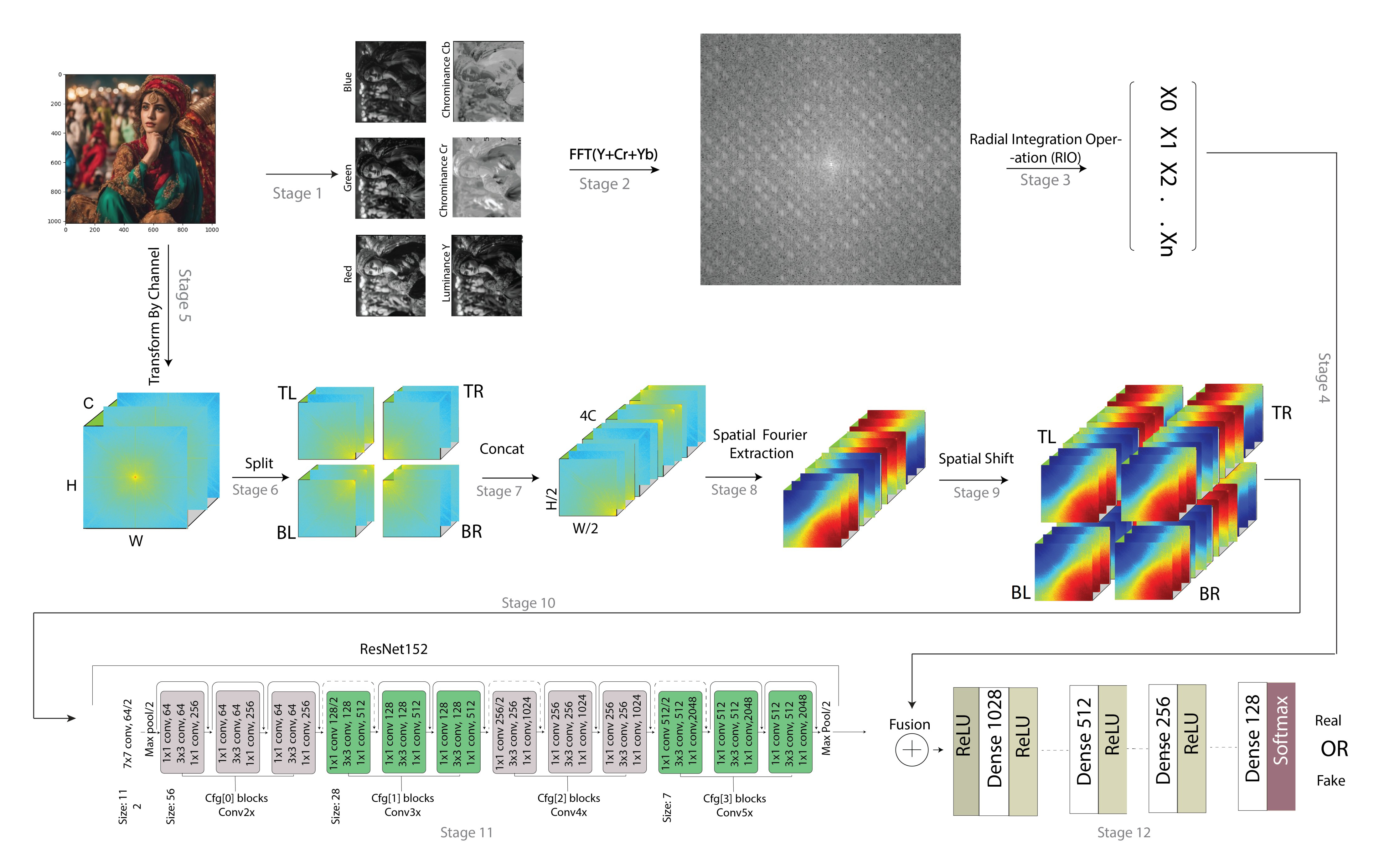}
  \caption{Overview of our approach, \SystemName: Stage 1 through 4 involves RGB to YCbCr conversion for luminance and chrominance extraction, followed by FFT for spectral analysis and Radial Integral Operation (RIO). And, Stage 5 through 10 show Spatial Fourier Unit (SFU) processes including splitting, concatenation, Spatial Feature Extraction (SFE), batch normalization, and spatial shifting as a multi-modal architecture. The input image is represented in 3D dimensions of height $(H)$, width $(W)$, and channels $(C)$. Finally, Stage 12 is the fusion of RIO from Stage 4 with ResNet architecture from Stage 11.}
  \label{fig:fm}
\end{figure*}

In this work, we propose~\SystemName, which combines spectral forensic analysis with deep learning classification to detect the latest AI-generated fake images effectively. The overview of our approach is presented in Fig.~\ref{fig:fm}. As shown in Fig.~\ref{fig:fm}, we start by pre-processing images through two parallel processes. First, we convert RGB images to YCbCr color space and apply FFT on each pixel, merging them into a $2D$ image represented in an $XY$ plane. Next, a Radial Integral Operation (RIO) operation is then applied to create a $1D$ array of vector values for different radii, capitalizing on the distinct power spectra of real and AI-generated images. The RIO output for fake AI-generated images is constant, while the fluctuations in the real images are not. 
In the next phase, we transform RGB images per channel and merge them to form a $3D$ image. FFT is applied to each channel, split, and concatenated into one image. We then employ Spatial Feature Extraction (SFE) to extract prominent features, followed by a spatial shift transformation. The processed image is fed into a ResNet152 architecture, culminating in softmax classification. In summary, our contributions include:
\vspace{-4pt}
\begin{itemize}
  \item 
  We introduce a novel Radial Integral Operation (RIO) in YCbCr color space, enhancing object recognition in diverse lighting components. 
  
  \item We propose a new Spatial Fourier Extraction (SFE) method to convert spatial features into the spectral domain, and globally update spectral data. 
  
  \item Our approach was rigorously tested with images from the latest AI-generative methods such as faces, scenes, and objects, and outperformed existing methods. 
\end{itemize}

\section{Related Work}
\label{sec:2}
\vspace{-2pt}

Several deepfake detection methods have been proposed in the past~\cite{woo1,woo2,woo3,woo4,woo5,woo6,woo7,woo8,woo9,woo10,woo11,woo12,woo13,woo14,woo15,woo16,woo17,woo18,woo19,woo20,woo21,woo22,woo23,woo24,woo25,woo26,woo27,woo28,woo29,woo30,woo31,woo32,woo33,woo34,Khalid2021FakeAVCelebAN}.
Wu et al. \cite{wu2023generalizable} developed a general classification approach, while Corvi et al. \cite{corvi2023intriguing} focused on pre-processing techniques to extract power spectra for improved classification. Radford et al. \cite{radford2021learning} proposed language supervision-based perceptual learning, and Wang et al. \cite{wang2023dire} introduced DIRE for detecting diffusion-generated images. Cozzolino et al. \cite{cozzolino2021universal} developed a GAN image detection method using a ResNet152 backbone, and Zhang et al. \cite{10221905} achieved high accuracy using Discrete Cosine Transform (DCT) for power spectra analysis. However, their focus was limited to a single stable diffusion dataset. Jeong et al. \cite{frepgan} use power spectrum and train the generated models on a generator and discriminator network. At the same time, we have applied RIO to accumulate the density of the power spectrum. In every radius, the density of the power spectrum for the fake images is mostly constant. We aim to use this phenomenon to classify real and fake images effectively. Other Forensic techniques have also been explored, with Corvi et al. \cite{Corvi2022_on} employing noise prints for camera fingerprint extraction. Mandelli et al. \cite{Mandelli_2022} used a forensic approach for distinguishing real and fake Western blot images, albeit limited to a specific image type. Ma et al. \cite{ma2023exposing} applied statistical and neural network-based methods to detect fingerprints in real vs. fake images, exploiting unique properties of the image generation process. However, our work differs from the above approaches because we have extracted fake footprints and distinct features in the frequency domain to more accurately and effectively classify the model fake methods.


\vspace{-5pt}
\section{Our Approach}
\label{sec:3}

\subsection{Radial Integral Operation (RIO)}
\label{ROI}
The initial stage of our method,~\SystemName~is designed to pre-process input RGB images to extract critical spectral information, where a dataset comprising inputs as real $X$ and AI-generated images $Y$ are defined as follows:
\begin{align}
\small
X \times Y = \{ (x_i, y_j) \mid x_i \in X, y_j \in Y \text{and } i \in \{1, \ldots,M\} \},
\end{align}

\vspace{-4pt}
\noindent where each $(x,y)$ is a pair of images containing real and fake images within our pre-processing pipeline, and M is the last index. 


\noindent \textbf{Conversion to YCbCr Color Space. }To prepare the images for spectral analysis, our initial step involves the conversion of input RGB images into the YCbCr color space. 
This transformation provides distinct channels for Luminance ($Y$) and Chrominance ($Cb$ and $Cr$). The Y channel, in particular, encapsulates essential image details, which are later used in extracting FFT features. 
Stage 1 in Fig. \ref{fig:fm} presents the conversion of RGB images to YCbCr images, while constants in the below equations are used to extract Luminance and Chrominance. The constants in the RGB to YCbCr conversion Eq. \ref{ycbcr} originate from the ITU-defined transformation matrix, reflecting perceived luminance and chrominance. They ensure accurate representation of brightness and color information in the YCbCr color space.
Specifically, the conversion equations are defined as follows:
{
\begin{equation}
\begin{split}
\label{ycbcr}
\small
    Y(i, j) &= 0.299 \cdot R(i, j) + 0.587 \cdot G(i, j) + 0.114 \cdot B(i, j) \\
    C_b(i, j) &= 128 - 0.168736 \cdot R(i, j) - 0.331264 \cdot G(i, j) \\
             &\quad + 0.5 \cdot B(i, j)  \\
    C_r(i, j) &= 128 + 0.5 \cdot R(i, j) - 0.418688 \cdot G(i, j)\\
             &\quad - 0.081312 \cdot B(i, j),
\end{split}
\end{equation}
}

\noindent where $Y(i,j)$ represents the luminance ($Y$) component of the YCbCr image, $Cb(i,j)$ corresponds to the blue chrominance ($Cb$) component, and $Cr(i,j)$ denotes the red chrominance ($Cr$) component, respectively. Furthermore, $R(i,j)$, $G(i,j)$, and $B(i,j)$ represent the red, green, and blue channel values of the pixel at position $(i,j)$ within the original RGB image, respectively.

\noindent \textbf{Fast Fourier Transformation (FFT). } Our next step involves applying the FFT operation to each pixel of the YCbCr image. We choose FFT, because the frequency information can effectively extract patterns for AI-generated fake methods. 
The FFT operation is defined as follows: $FFT(k) = \sum_{n=0}^{N-1} ( \text{Re}(x(n)) + j \cdot \text{Im}(x(n)) ) e^{-j\frac{2\pi}{N}kn}$. And, it is applied to each channel of the YCbCr in the following way: $Y'(i, j) = \text{FFT}(Y(i, j))$, $Cb'(i, j) = \text{FFT}(Cb(i, j))$, and $Cr'(i, j) = \text{FFT}(Cr(i, j))$.The outcome of the FFT operation encompasses both real ($Re$) and imaginary ($Im$) components for each pixel, effectively transforming the image into the frequency domain $FFT(k)$ on the basis of complex discrete-time signals $x(n)$.

\noindent \textbf{Merging Spectral Information. }The real $Re$ and imaginary $Im$ components derived from all channels are subsequently merged into a singular 2D image, where Stage 2 in Fig. \ref{fig:fm} characterizes the result from the operations. This combined image serves as a representation of spectral information within the XY  coordinates and is mathematically expressed as follows:
{
\begin{equation}
\small
    XY(i, j) =  \text{FFT}(Y(i, j)) +  \text{FFT}(Cb(i, j)) + \text{FFT}(Cr(i, j))
\end{equation}
}

\subsubsection{Quadrant Analysis and Radial Integral Operation (RIO). }

To further enhance the valuable features from spectral content, we apply an RIO, which draws a circle across the  2D plane from each radius of the image spanning from (0,0) point and employs an integration function to compute spectral information at varying radii. The simplified radial integral operation (RIO) in Stage 3 in Fig. \ref{fig:fm} is defined as follows:
\begin{equation}
\label{simp_RIO}
\small
    \bar{f}(r) = \frac{1}{2\pi} \int_{0}^{2\pi} f(r, \theta) \, d\theta
\end{equation}
This RIO culminates in a 1D array capturing RIO value across multiple radii, encompassing numerous pixels. And, finding all pixel's Fourier weight within the individual radius is defined as follows:
\begin{equation}
\small
\mathbb{M} = \sum \sum |F(u,v)|,
\end{equation}
\noindent where $F(u,v)$ represents the Fourier coefficient at spatial frequency coordinates ($u$,$v$) and $\sum \sum$ is taken over the Fourier co-efficient with the circular region radius $R_i$, and ($u$,$v$) needs to satisfy the following condition:
\begin{equation}
\small
(u-u_\text{center})^2 + (v-v_\text{center})^2 <= R^2
\end{equation}

From Eq.~\ref{simp_RIO}, our expected $f(r, \theta)$ can be derived by computing the integral of the squared magnitude of the Fourier transform of the image over a range of radii $r_i$. This operation captures the spatial frequency distribution of the image, enabling analysis of its structural characteristics and frequency components. 
\begin{equation}
\small
    f(r, \theta)  = || \left(\int\limits_{0}^{R_i} \left(\frac{3\mathbb{M}}{4\pi r_i^3}\right)r_i dr_i\right) \left(w_k \cos(\theta), w_k \sin(\theta)\right) ||^2,
\end{equation}
where $R_i=r_1,r_2,...r_n$ and $W_k=\frac{2 \pi k}{N}$, and $k$ is the magnitude value, and $N$ is the total number of pixels on that $r_i$, respectively. In Section~\ref{Spec-Ana-RIO}, we present that, for different fake images, the spectrum lines in RIO do not overlap each other. The advantage of non-overlapping frequency information is that fake images fluctuate constantly, which differs from real images. This indicates and captures that RIO provides valuable information and can assist in representing features to distinguish real vs. fake images.

\subsection{Spatial Fourier Unit (SFU)}
\label{Spatial Fourier Extraction}

Next to extract the spatial features from the frequency domain of the images, which involves the following steps: First, let $I_{\text{RGB}}(i, j, c)$ represent the input RGB image for each color channel R, G, and B, and let $I_{\text{FFT}}(i, j, c)$ be the image in the frequency domain after FFT, where each channel from RGB individually transforms with FFT in the following way:
\begin{equation}
\small
I_{\text{FFT}}(i, j, c) \rightarrow \text{FFT}(I_{\text{RGB}}(i, j, c)) 
\end{equation}
After that, we split a Fourier transformed image $I_{\text{FFT}}(i,j,c)$ into four quadrants as shown in Eq.~\ref{eq:split-fft}: Top Left (TL), Top Right (TR), Bottom Left (BL), and Bottom Right (BR). This split operation in Eq.~\ref{eq:split-fft} plays a pivotal role in extracting essential spatial frequency features for Spatial Fourier Extraction (SFE), which halves both spatial dimensions and renders four smaller feature maps for retaining valuable artifacts (See Section~\ref{effect} for more explanation).
\setlength{\parskip}{0pt}
\begin{equation}
\label{eq:split-fft}
\small
\begin{aligned}
I_{\text{split}} &\rightarrow \text{SPLIT}(I_{\text{FFT}}(i, j, c)) \\
I_{\text{split}_{\text{TL}}} &= I_{\text{FFT}}(i,j) \text{ for } 1 \leq i \leq \frac{H}{2}, 1 \leq j \leq \frac{W}{2} \\
I_{\text{split}_{\text{TR}}} &= I_{\text{FFT}}(i,j+\frac{W}{2}) \text{ for } 1 \leq i \leq \frac{H}{2}, \frac{W}{2} + 1 \leq j \leq W \\
I_{\text{split}_{\text{BL}}} &= I_{\text{FFT}}(i+\frac{H}{2},j) \text{ for } \frac{H}{2} + 1 \leq i \leq H, 1 \leq j \leq \frac{W}{2} \\
I_{\text{split}_{\text{BR}}} &= I_{\text{FFT}}(i+\frac{H}{2},j+\frac{W}{2}) \text{ for } \frac{H}{2} + 1 \leq i \leq H, \\
             &\quad \frac{W}{2} + 1 \leq j \leq W,
\end{aligned}
\end{equation}

\noindent where the $SPLIT$ function performs the split operation in four parts: \(I_{\text{split}_{\text{TL}}}\), \(I_{\text{split}_{\text{TR}}}\), \(I_{\text{split}_{\text{BL}}}\), and \(I_{\text{split}_{\text{BR}}}\) as shown in Stage 6 in Fig. \ref{fig:fm}, and, the channel $C$ represents the number of channels in each of the splitting parts.

Also, we define \(I_{Concat}\) to be a new image formed by concatenating the above four parts along the channel dimension in Eq.~\ref{eq:concat}, where \(I_{Concat}\) stacked feature maps from Eq.~\ref{eq:split-fft} for increasing nonlinearities, And, \(I_{Concat}\) constrcuts deeper layers, which is less prone to overfitting, as they can capture more diverse and discriminative features. The new image, \(I_{Concat}\) has \(4C\) channels depicts in Stage 7 in Fig. \ref{fig:fm}. This operation results in an image with the height and width (\(H/2\) and \(W/2\)), respectively, with an increased number of channels, combining information from all four parts into a single image with richer channel-wise information as shown in Stage 7.
\begin{equation}
\label{eq:concat}
\small
\begin{aligned}
    I_{Concat} &= \text{Concat}([I_{\text{split}_{\text{TL}}}, I_{\text{split}_{\text{TR}}}, I_{\text{split}_{\text{BL}}}, I_{\text{split}_{\text{BR}}}], \text{axis}=2)
\end{aligned}
\end{equation}
Furthermore, we apply the Spatial Fourier Extraction (SFE) operation in Stage 8 as shown in Fig.\ref{fig:fm}. The SFE transforms spatial features into a frequency domain, conducting efficient global updates on frequency data. The SFE is used in Eq.~\ref{eq:latent} to obtain the $I_{Latent}$, where  $I_{Latent}$ contains all the important spectral feature information from $I_{\text{Concat}}$. Deriving $I_{Latent}$ from Eq.~\ref{eq:latent} is further discussed in Section~\ref{SFE-det}.
\begin{equation}
\label{eq:latent}
\small
\begin{aligned}
I_{\text{Latent}}(i, j, c) \rightarrow \text{Spatial\_Fourier\_Extraction}(I_{\text{Concat}}(i, j, c))
\end{aligned}
\end{equation}
\subsubsection*{Spatial Fourier Extraction (SFE)}
\label{SFE-det}
Our proposed method SFE in Stage 8 is shown in Figure~\ref{fig:fm}, which extracts spectral features from the augmented \(I_{Concat}\) image. The SFE is conducted in the following three steps:

\noindent \textbf{Step 1.} The spatial operation applies a depthwise convolution kernel $K$ to a feature map $X$ as follows:
{ 
\begin{equation}
\label{Spatial}
\small
    Y_{i,j}^{(k)} = \sum_{l=1}^{\mathbf{C}} X_{i+m,j+n}^{(l)} \ast K_{m,n}^{(l)},
\end{equation}
}

\noindent where \(i\) and \(j\) represent spatial indices, and $Y_{i,j}(k)$ represents the output feature map at position $(i,j)$ with depth $k$, and
$X_{i+m,j+n(l)}$ represents the input feature map at position $(i+m,j+n)$ with depth $l$. And, $K_{m,n(l)}$ represents the depthwise convolution kernel at position $(m,n)$ with depth $l$, and $\mathbf{C}$ represents the augmented height from Eq.~\ref{eq:concat}, and the symbol $\ast$ denotes the convolution operation.

\noindent \textbf{Step 2.} The next step of our SFE method is to normalize the output from Eq.~\ref{Spatial}. First, let us define that we have a feature map ($Y$) after the previous step with a shape $(N, \mathbf{C}, H, W)$, where $N$ is the batch size, $\mathbf{C}$ is the number of channels, and $H$ and $W$ are the spatial dimension, respectively. 


Next, we introduce a covariance-based normalization approach to calculate the second-order statistics (covariance matrix) $Cov_\mathbf{c}$ within each channel $\mathbf{c}$ for each mini-batch as follows: 
\begin{equation}
    \label{2ndNorm:1}
        Cov_\mathbf{c} = \frac{1}{N} \sum_{i=1}^{N} (Y[i, \mathbf{c}, :, :] - \mu_\mathbf{c})(Y[i, \mathbf{c}, :, :] - \mu_\mathbf{c})^T,
\end{equation}
\noindent where $\mu_\mathbf{c}$ represents the batch mean for channel $\mathbf{c}$. For each channel $\mathbf{c}$, we normalize the activations using the inverse square root of the covariance matrix to stabilize the operation as follows:
\begin{equation}
    \label{2ndNorm:2}
    \small
        Z[i, \mathbf{c}, :, :] = \gamma_\mathbf{c} (Y[i, \mathbf{c}, :, :] - \mu_\mathbf{c}) (Cov_\mathbf{c} + \epsilon)^{-1/2} + \beta_\mathbf{c},
\end{equation}
where \noindent $\gamma$ is the shift and $\beta$ is the scale parameter, and $\epsilon$ is a small constant added for numerical stability, which is in the range of $\epsilon$ and is typically in the order of magnitude of $1 \times 10^{-5}$ to $1 \times 10^{-6}$. This range is small enough to avoid numerical instability while not affecting the normalization process significantly. This covariance normalization in Eq.~\ref{2ndNorm:2} stabilizes the activations and standardizes the feature distributions across channels and spatial locations. Specifically, subtracting the channel mean centers the data while normalizing by the square root of the covariance matrix scales the variances. The learnable parameters $\gamma_\mathbf{c}$ and $\beta_\mathbf{c}$ further tune the normalized activations.

Overall, this data-dependent normalization in Step 2 adaptively standardizes the representations spatially and across channels. We can model more complex second-order statistics for effective feature normalization than simple channel-wise means and variances. The normalized output $Z$ can also serve as more robust intermediate representations for subsequent processing in Step 3.
 

\noindent \textbf{Step 3.} The last step in SFE is to apply a modified ReLU activation function is applied after the depthwise normalization process, which can be represented as follows:
\begin{equation}
\label{Relu:1}
\small
\begin{aligned}
    \text{ReLU}(Z) &= \max(0, Z); \\
    \text{FReLU}(Z) &= \text{ReLU}(\text{Re}(Z)) + i\text{ReLU}(\text{Im}(Z)),   
\end{aligned}
\end{equation}
\noindent where \(Z\) is an element in the feature map. Assuming that the positive and negative values of the complex-valued images are represented in the four quadrants, \(\text{FReLU}\) has the advantage that information can be obtained from three quadrants among four. Applying the inverse FFT on both real and imaginary components, Eq. \ref{Relu:1} guarantees the generation of a Hermitian matrix characterized by its symmetry. It facilitates the production of real-valued outputs suitable for subsequent neural network computations~\cite{Bergland}. 

\vspace{10pt}
\noindent Lastly, in SFU, let \(I_1(i, j, c)\), \(I_2(i, j, c)\), \(I_3(i, j, c)\), and \(I_4(i, j, c)\) be the four copies of the \(I_{Latent}\). And, we stack them in the specific arrangement via a spatial shift method as follows:
\begin{align*}
\small
I_{\text{Spatial\_Shift}}(i, j, c) =
\begin{cases}
I_1(i, j, c) \\
I_2(i, j + \frac{W}{2}, c - 4C) \\
I_3(i + \frac{H}{2}, j, c - 8C) \\
I_4(i + \frac{H}{2}, j + \frac{W}{2}, c - 12C), 
\end{cases}
\end{align*}
where
{
\begin{equation}
\label{eq:spatial_shift}
\small
\begin{aligned}
I_1 & \in \{(i, j, c) \mid 1 \leq i \leq \frac{H}{2}, 1 \leq j \leq \frac{W}{2}, 1 \leq c \leq 4C\}, \\
I_2 & \in \{(i, j, c) \mid 1 \leq i \leq \frac{H}{2}, \frac{W}{2} + 1 \leq j \leq W, 4C + 1 \leq c \leq 8C\}, \\
I_3 & \in \{(i, j, c) \mid \frac{H}{2} + 1 \leq i \leq H, 1 \leq j \leq \frac{W}{2}, 8C + 1 \leq c \leq 12C\}, \\
I_4 & \in \{(i, j, c) \mid \frac{H}{2} + 1 \leq i \leq H,  \frac{W}{2} + 1 \leq j \leq W, \\
& 12C + 1 \leq c \leq 16C\},
\end{aligned}
\end{equation}
}
In the above representation, \(I_{\text{Spatial\_Shift}}\) is the stacked image created by arranging \(I_1\), \(I_2\), \(I_3\), and \(I_4\) in a specific manner: 
\(I_1\) in the top-left, \(I_2\) in the top-right, \(I_3\) in the bottom-left, and \(I_4\) in the bottom-right positions, as shown in Stage 9 in Fig. \ref{fig:fm}. The height and width dimensions can be perfectly restored to the original size before any downsampling or manipulations by generating four shift-copied versions of each activation in this precise layout. Propagating this quadruplicated information facilitates lossless transmission of all visual details encoded within the initial feature map. Irrecoverable losses can occur due to misalignment between original and transformed feature grid dimensions without this tailored spatial shifting procedure to copy and rearrange components. Overall, this deliberate copying and positioning enables dimensional restoration without permanent losses of spatial information. Further explanation for the significance of split, concat, and shift is demonstrated in Section~\ref{effect}.

\section{Experimental Results and Analysis}
\label{sec:4}
\begin{figure}[htbp]                             
  \centering
  \includegraphics[width=\linewidth]{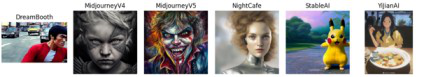}
  \caption{Each row displays images generated by Diffusion Models (DM) sourced from various online platforms.}
  \label{fig:figurepractical}
\end{figure}

\noindent \textbf{Datasets. }Our dataset consists of both existing popular datasets from Wu et al. \cite{wu2023generalizable} and self-generated images using the latest open source methods such as Stable diffusionV1.2~\cite{rombach2021highresolution} by StableAI~\cite{stabilityai2022}, DreamBooth~\cite{dreambooth2022}, and Latent Diffusion by CompVis models. Fake images in the test dataset contain samples from eleven generation methods such as ProGAN \cite{karras2018progressive}, StyleGAN2 \cite{karras2020analyzing}, StyleGAN3 \cite{zhu2023stylegan3}, BigGAN \cite{brock2018large}, EG3D \cite{Chan2022}, Taming Transformer \cite{esser2021taming}, DALL-E 2 \cite{ramesh2022dalle}, GLIDE \cite{patashnik2021stylegannada}, Latent Diffusion \cite{ho2020denoising}, Guided Diffusion ~\cite{NEURIPS2021_49ad23d1} and Stable Diffusion v1.2. And, real test images are from ImageNet \cite{deng2009imagenet}, COCO \cite{lin2014microsoft}, and, Danbooru \cite{danbooru2021}. Additionally, we generate a practical test dataset $\mathbf{T_{Gen}}$, which has 1K  images for each category from DreamBooth, MidjourneyV4~\cite{midjourney2022}, MidjourneyV5~\cite{midjourney2022}, NightCafe~\cite{nightcafe2019}, StableAI, and YiJian~\cite{yijian}.
In particular, we include test images with significant variations, such as faces and objects, as well as differing indoor and outdoor environments, landscapes, and scenes. 
The total composition of the entire training dataset contains 430K authentic images and 410K fake images. We have taken 5K for testing from each generation method. 
Also, we have used many prompts to generate images using stable diffusion V1.2, midjourney V5, Dreambooth, etc. Here are some samples of the prompts that we used for image generation in Fig~\ref{fig:figurepractical} : Bruce Lee sitting in a car on a road way, Pikachu standing on roadway, seaside, etc.

\begin{table*}[ht]
  \caption{Comparison with SOTA Methods: The first column represents a combination of the actual dataset, while the Generative Adversarial Network (GAN) and Diffusion Methods (DM) are presented in the subsequent columns. The last row, `Average,' represents the average performance of each method against all the combined datasets.
  Bold values are the best value, and underlined values are the second best. All of the SOTA methods are trained and compared with the same datasets.}
  \label{tab:table2} 
  \small
  \begin{tabular*}{\textwidth}{
  @{\extracolsep{\fill}}
  p{0.07\textwidth} 
  p{0.03\textwidth} 
  p{0.14\textwidth} 
  p{0.03\textwidth} 
  p{0.03\textwidth} 
  p{0.03\textwidth} 
  p{0.03\textwidth} 
  p{0.03\textwidth} 
  p{0.03\textwidth} 
  p{0.03\textwidth} 
  p{0.03\textwidth} 
  p{0.03\textwidth} 
  p{0.03\textwidth} 
  p{0.03\textwidth} 
  p{0.03\textwidth} 
  @{\extracolsep{\fill}}
  }
  \hline
  \shortstack{Real\\Dataset} & \multicolumn{2}{c}{\shortstack{Fake\\Dataset}} & \multicolumn{2}{c}{Grag} & \multicolumn{2}{c}{CR} & \multicolumn{2}{c}{Wang} & \multicolumn{2}{c}{Zhang} & \multicolumn{2}{c}{Chai} & \multicolumn{2}{c}{Ours} \\

  \cmidrule(lr){2-3} \cmidrule(lr){4-5} \cmidrule(lr){6-7} \cmidrule(lr){8-9} \cmidrule(lr){10-11} \cmidrule(lr){12-13} \cmidrule(lr){14-15}
  & & & AUC & Acc & AUC & Acc & AUC & Acc & AUC & Acc  & AUC & Acc & AUC & Acc\\

  \hline

  \multirow{11}{*}{\shortstack{ImageNet\\COCO}} & \multirow{4}{*}{GAN} & BigGAN &0.745 &\underline{0.796} &0.725 &0.534 &\underline{0.858} &0.795 &0.485 &0.497 &0.653 &0.504 & \textbf{0.951} & \textbf{0.936} \\
  & & StyleGAN2 &0.858 &\underline{0.912}  &0.870 &0.558 &\underline{0.899} &0.728 &0.505 &0.518 &0.736 &0.508 & \textbf{0.967} & \textbf{0.928}\\
  & & StyleGAN3 &\underline{0.908} &\underline{0.854} &0.869 &0.615 &0.901 &0.754 &0.519 &0.529 &0.767 &0.502 & \textbf{0.921} & \textbf{0.877}\\
  & & ProGAN &0.833 &0.772 &0.794 &0.654 &\underline{0.880} &\underline{0.815} &0.485 &0.497 &0.653 &0.504 & \textbf{0.987} & \textbf{0.957}\\
  & & EG3D &0.793 &0.668 &0.856 &0.537 &\underline{0.860} &\underline{0.799} &0.606 &0.589 &0.819 &0.498 & \textbf{0.872} & \textbf{0.834}\\
 
  & \multirow{7}{*}{DM} & DALL-E 2 &0.516 &0.552 &0.522 &0.520 &0.586 &0.560 &\underline{0.650} &\underline{0.620} &0.584 &0.497 & \textbf{0.941} & \textbf{0.872}\\
  & & GLIDE &0.574 &0.588 &0.624 &0.528 &0.608 &\underline{0.600} &0.525 &0.531 &\underline{0.715} &0.510 & \textbf{0.936} & \textbf{0.927}\\

  & & \text{Latent Diffusion} &\underline{0.863} &0.675 &0.844 &\underline{0.907} &0.749 &0.650 &0.463 &0.479 &0.652 &0.506 &\textbf{0.970} &\textbf{0.921}\\
  & & \text{Taming Transformer} &0.710  &0.692 &0.757 &0.703 &\underline{0.943} &0.652 &0.791 &\underline{0.790}  &0.741 &0.610 &\textbf{0.950} &\textbf{0.876}\\
  
  & & \text{Stable DiffusionV1.2} &0.610 &0.600 &0.598 &0.563 &0.587 &\underline{0.621} &0.445 &0.465 &\underline{0.772} &0.511 &\textbf{0.972} &\textbf{0.942}\\
  & & \text{Guided Diffusion} &0.588 &0.577 &0.584 &0.520 &0.566 &\underline{0.652} &.491 &0.491 &\underline{0.691} &0.510 &\textbf{0.925} &\textbf{0.915}\\
  \hline
  
  \multirow{11}{*}{\shortstack{Artist\\Danbooru}} & \multirow{5}{*}{GAN} & BigGAN &\textbf{0.949} &0.883 &0.892 &0.958 &0.946 &\textbf{0.984} &0.705 &0.733 &0.857 &0.725 & \underline{0.942} & \underline{0.956}\\
  & & StyleGAN2  &0.951 &0.883 &0.911 &0.899 & \underline{0.969} & \underline{0.972} &0.770 &0.713 &.883 &0.702 &\textbf{0.985} &\textbf{0.982}\\
  & & StyleGAN3 &\underline{0.978} &0.951 &0.966 &0.967 &0.969 &\textbf{0.972} &0.440 &0.338  &0.718 &0.500 &\textbf{0.983} &\underline{0.970}\\
  & & ProGAN &0.970 &0.899 &0.935 &\underline{0.986} &0.952 &0.991 &\underline{0.992} &0.835 &0.976 &0.649 & \textbf{0.992} & \textbf{0.992}\\
  & & EG3D &0.823 &0.722 &0.826 &0.787 &\underline{0.905} &\underline{0.943} &0.803 &0.896 &0.500 &0.784 & \textbf{0.962} & \textbf{0.985}\\
  & \multirow{6}{*}{DM} & DALL-E 2 &0.778 &0.616 &0.782 &\underline{0.827} &0.746 &0.737 &\underline{0.933} &0.821 &0.525 &0.500 &\textbf{0.955} &\textbf{0.898}\\
  & & GLIDE &0.827 &0.668 &0.754 &\underline{0.845} &0.754 &0.814 &0.819 &0.734 &\underline{0.966} &0.503 &\textbf{0.974} &\textbf{0.957}\\
  & & \text{Stable DiffusionV1.2} &0.795 &0.689 &0.788 &0.745 &\underline{0.929} &\underline{0.871} &0.777 &0.718 &0.850 &0.698 &\textbf{0.982} &\textbf{0.976}\\
  & & \text{Guided Diffusion} &\underline{0.869} &0.670 &0.818 &\underline{0.897} &0.716 &0.681 &0.793  &0.769 &0.831 &0.644 &\textbf{0.988} &\textbf{0.974}\\
  & & \text{Latent Diffusion} &\underline{0.863} &0.675 &0.844 &\underline{0.908} &0.749 &0.650 &0.785 &0.721 &0.843 &0.678 &\textbf{0.971} &\textbf{0.959}\\
  & & \text{Taming Transformer} &0.865 &0.651 &\underline{0.878} &\underline{0.931} &0.966 &0.915 &0.866 &0.796 &0.694 &0.545 &\textbf{0.969} &\textbf{0.960}\\

 \midrule
  
  \multicolumn{3}{c}{Average Performance with All Datasets} & 0.799 & 0.729 & 0.787 & 0.746 &\underline{0.815} &\underline{0.781} & 0.662 & 0.637  & 0.738 & 0.570 &\textbf{0.957} &\textbf{0.935}\\
  \hline
  \end{tabular*}
\end{table*}


\noindent \textbf{Experimental Settings.} 
We use PyTorch (ver. 3.6) on CUDA 11.7 with two Nvidia Titan RTX GPUs for experiments. A batch size of 256 is used, and all images are resized to 224 
$\times$ 224. The cross entropy loss function and Adam optimizer with a learning rate of 0.1. For evaluation metrics, AUC and accuracy are used. 


\subsection{Comparison with SOTA Methods}
We extensively evaluated our approach on different datasets, utilizing various pre-trained ResNet models for feature extraction. These models include ResNet18, ResNet32, ResNet50, ResNet101, and ResNet152, respectively. We show that ResNet152 is the most suitable choice for the ablation study. Hence, we use ResNet152 to evaluate other SOTA methods in Table~\ref{tab:table2}.
As shown in the last row in Table~\ref{tab:table2}, our method achieves the highest average AUC and accuracy, compared to the existing SOTA approaches: Wang et al., \cite{wang2023dire}, Chandrasegaran et al., \cite{chandrasegaran2022discovering}, Chai et al., \cite{patchforensics}, Grag et al., \cite{9428429}, and Xu Zhang et al.~\cite{zhang2019detecting}. We find that our approach outperforms other SOTA methods in the majority of cases, except BigGAN, which is the second best.
In addition, we conducted an experiment involving the testing of 10k images, as delineated in Table \ref{tab:table5}, and found that our proposed method performs better in classifying the real vs. fake in terms of average accuracy compared to CLIP on all methods.
\begin{table}[!ht]
\caption{Comparison of fake image detection with CLIP method vs. Ours, where accuracy is used for evaluation.}
\label{tab:table5}
\small
\begin{tabularx}{\linewidth}{>{\centering\arraybackslash}X>{\centering\arraybackslash}X>{\centering\arraybackslash}X}
\toprule
\textbf{Method} & \textbf{CLIP (\%)} & \textbf{Ours (\%)} \\
\midrule
ProGAN & 90.1 & \textbf{99.2} \\
StyleGAN2 & 83.0 & \textbf{98.2} \\
StyleGAN3 & 84.0 & \textbf{97.0} \\
GLIDE & 93.7 & \textbf{95.7} \\
Latent Diffusion & 89.9 & \textbf{95.9} \\
Stable DiffusionV1.2 & 72.9 & \textbf{97.6} \\
EG3D & 90.8 & \textbf{98.5} \\
BigGAN & 98.2 & \textbf{95.6} \\
DALL-E 2 & 51.1 & \textbf{89.8} \\ 
\midrule
\multicolumn{1}{c}{\textbf{Average}} & 83.74 & \textbf{96.40} \\
\bottomrule
\end{tabularx}
\end{table}

\vspace{-6pt}
\subsection{Ablation Studies}


\subsubsection{\textbf{Effect of Different ResNet architectures}} 

We analyzed the accuracy performance of various ResNet architectures (see Table \ref{tab:table3}).
Across all ResNet architectures, we consistently noticed increased performance with our SFU and RIO component.
Specifically, for ResNet50, mean accuracy improves from 67\% without additional components to 82.1\%, with RIO + SFU. Similarly, the performance of ResNet101 increases from 75.2\% to 87.2\%. ResNet152 showed increased performance from 78.7\% to 93.0\% . 

\subsubsection{\textbf{Effect of Our Proposed Components}}
\noindent \textbf{SFU} enhances spatial feature extraction with advanced up-scaling techniques. This SFU component resulted in high-resolution datasets and increased performance across all ResNet architectures. The most significant gains are observed in high-resolution datasets such as Artist \& Danbooru, illustrating SFU's effectiveness in enhancing spatial details. 
Also, \textbf{RIO} enhances the regions of interest, reducing computational overhead by deriving a 1D array from an image while increasing accuracy across all models, especially in ResNet152 combined with SFU. This results in 3\% increase. This shows RIO improves feature extraction for accurate classification.  
Moreover, \textbf{SFU + RIO} integration yields accuracy improvements, notably 3\% increment in ResNet152, which is effective in complex datasets such as Artist \& Danbooru. This improvement across different ResNet architectures indicates the robustness of the combined approach. 

Also, \textbf{SFU without Split and Shift ($\widehat{SFU}$)} shows decreased accuracy across all architectures, with ResNet50 dropping from 77.6\% to 72.42\% and ResNet152 from 90\% to 84.82\%. The absence of split and shift leads to less effective feature extraction. Lastly, \textbf{RIO without YCbCr Conversion ($\widehat{RIO}$)} results in reduced accuracy for ResNet50 from 82.1\% to 79.72\%, which is more pronounced in color-critical datasets such as AD/GLIDE and AD/LD. ResNet152's accuracy falls from 93\% to 91.58\%, affirming the importance of YCbCr conversion in maintaining high performance across architectures.


\begin{table}[!ht]
\caption{Ablation Studies: Assessing the effects of different ResNet architectures and SFU and RIO components over various fake image generation methods, where $\widehat{SFU}$ represents SFU w/o split and w/o shift, and $\widehat{RIO}$ indicates RIO w/o YCbCr, respectively.}
\label{tab:table3}
\scriptsize
\small
\setlength{\tabcolsep}{1pt} 
\resizebox{\columnwidth}{!}{ 
\begin{tabular}{cccccccc}
\toprule
\multirow{2}{*}{Architecture} & \multirow{2}{*}{Components} & \multicolumn{5}{c}{Test Datasets (Real/Fake)-Accuracy} \\
\cmidrule(lr){3-7} 
 &  & IM/SG2 & IM/DALL-E 2 & AD/GLIDE & AD/LD & Average \\
\midrule
ResNet50 & None & 0.693 & 0.659 & 0.677 & 0.661 &0.670 \\
ResNet50 & SFU & 0.779 & 0.768 & 0.764 & 0.773 & 0.776 \\
ResNet50 & $\widehat{SFU}$ + RIO &0.739 & 0.712 & 0.727 & 0.717 & 0.7242 \\
ResNet50 & $\widehat{RIO}$ + SFU & 0.817 & 0.772 & 0.794 & 0.794 & 0.7972 \\
ResNet50 & RIO + SFU & \textbf{0.831} & \textbf{0.787} & \textbf{0.816} & \textbf{0.834} & \textbf{0.821} \\
\midrule
ResNet101 & None & 0.766 & 0.721 & 0.762 & 0.734 & .752 \\
ResNet101 & SFU & 0.851 & 0.796 &0.853 & 0.867  & 0.845 \\
ResNet101 & $\widehat{SFU}$ + RIO & 0.790 & 0.745 & 0.796 & 0.791 & 0.790 \\
ResNet101 & $\widehat{RIO}$ + SFU & 0.865 & 0.812 & 0.871 &0.879 & 0.857 \\
ResNet101 & RIO + SFU & \textbf{0.871} & \textbf{0.818} & \textbf{0.899} & \textbf{0.882} & \textbf{0.872} \\
\midrule
ResNet152 & None & 0.817 & 0.764 & 0.786 & 0.772 & 0.787 \\
ResNet152 & SFU & 0.903 & 0.843 & 0.922 & 0.916  & 0.900 \\
ResNet152 & $\widehat{SFU}$ + RIO & 0.856 & 0.809 & 0.864 & 0.858  & 0.848 \\
ResNet152 & $\widehat{RIO}$ + SFU & 0.922 & 0.868 & 0.941 & 0.947 & 0.9158 \\
ResNet152 & RIO + SFU & \textbf{0.928} & \textbf{0.872} & \textbf{0.957} & \textbf{0.959} & \textbf{0.930} \\
\bottomrule
\end{tabular}
} 
\end{table}

\subsubsection{\textbf{Effectiveness of Split and Shift Operation in SFU}} 
\label{effect}

Incorporating shifting operations is essential for efficient information exchange among neighboring pixels. The shift operation strategically replicates critical features across top-right, top-left, bottom-right, and bottom-left positions, ensuring the retention of crucial information when feeding SFU-extracted features into the ResNet architecture and mitigating information loss. As assessed by accuracy (Acc), in Table \ref{tab:table4}, we present the outcomes delineating the efficacy of split and shift operations on the practical dataset $\mathbf{T_{gen}}$. We can observe that the preeminent methods. It is noteworthy that all methodologies were trained on the ProGAN subset and subsequently assessed for generalization across the remaining 6 subsets. The variants denoted as Wang-$0.1$ and Wang-$0.5$ signify models trained with 10\% and 50\% data augmentation, respectively. 

Our proposed methodology, designated as ``Ours'', attains the highest accuracy across all test datasets, achieving \textbf{0.936} for DreamBooth \cite{dreambooth2022}, \textbf{0.899} \cite{midjourney2022} for MidjourneyV4, \underline{0.853} for MidjourneyV5 \cite{midjourney2022}, \textbf{0.922} for NightCafe \cite{nightcafe2019}, \textbf{0.878} for StableAI \cite{stabilityai2022}, and \underline{0.731} for YiJian \cite{yijian}. These results demonstrate our approach's efficacy in enhancing performance on practical datasets with split and shift operations.

 \begin{table}[!ht]
\caption{Effect of split and shift operation on practical dataset $\mathbf{T_{Gen}}$. We use accuracy for evaluation, where the best values are shown in bold, and the second best are underlined. Noted that all the methods are trained with our dataset while being tested for generalization on the six subsets. Wang-$0.1$ and Wang-$0.5$ represent two variants trained with 10\% and 50\% data augmentation, respectively.}
\label{tab:table4}
\small
\setlength{\tabcolsep}{2pt} 
\resizebox{\columnwidth}{!}{
\begin{tabular}{cccccccc}
\toprule
\multirow{2}{*}{Method} & \multicolumn{6}{c}{$\mathbf{T_{gen}}$-Accuracy} & \multirow{2}{*}{Average} \\
\cmidrule(lr{1em}){2-7} 
&  \shortstack{Dream\\Booth} &  \shortstack{Midjourney\\V4} &  \shortstack{Midjourney\\V5} &  \shortstack{Night\\Cafe} &  \shortstack{Stable\\AI}  &  \shortstack{Yi\\Jian} \\
\midrule
Wang-$0.5$ & 0.849 & 0.839 & 0.805 & 0.821 & 0.810 & \textbf{0.739} & 0.811 \\
Wang-$0.1$ & \underline{0.857} & \underline{0.871} & \textbf{0.891} & \underline{0.877} & \underline{0.842} & 0.725 & \underline{0.843} \\
CR & 0.774 & 0.761 & 0.735 & 0.800 & 0.780 & 0.592 & 0.741 \\
Grag & 0.665 & 0.692 & 0.675 & 0.741 & 0.739 & 0.547 & 0.677 \\
\midrule
 \shortstack{Ours-w/o split \\ \&  w/o shift}  & 0.845 &0.783 & 0.749 & 0.805 & 0.757 & 0.711 & 0.775 \\
Ours-w/o shift & 0.853 & 0.834 &0.771 & 0.821 & 0.836 & \underline{.794} & 0.812 \\
Ours  & \textbf{0.936} & \textbf{0.899} & \underline{0.853} & \textbf{0.922} & \textbf{0.878} & \underline{.731} & \textbf{0.870} \\
\bottomrule
\end{tabular}
}
\end{table}


\vspace{-2pt}
\subsubsection{\textbf{Effectiveness of Data Augmentation}} 

\begin{figure}[htbp]
  \centering
  \includegraphics[width=0.7\columnwidth]{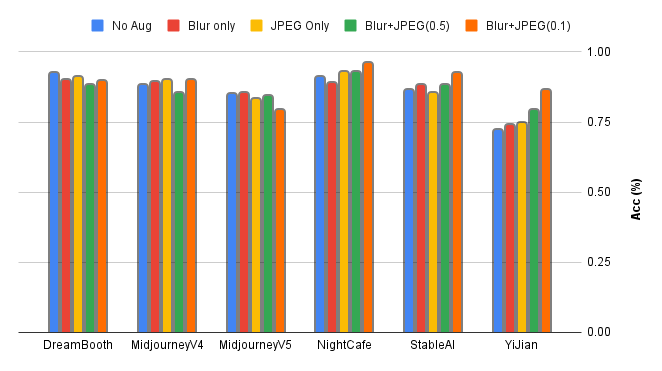}
  \caption{The efficacy of augmentation methods on detector performance is evaluated. All detectors are trained using ProGAN and assessed on alternative generators, with their respective accuracies presented. Augmentation generally enhances performance, although noteworthy exceptions, such as MidjourneyV5, are observed.}
  \label{fig:figureaug}
  
\vspace{-10pt} 
\end{figure}

Data Augmentation evaluates how well each detector handles post-processing on images with various adjustments. We consider four operations based on \cite{wang2020cnngenerated}: (1) No augmentation; (2) Gaussian blur with a 50\% chance, blur strength (sigma) from 0 to 3; (3) JPEG compression with a 50\% chance, compression quality from 30 to 100; (4a) Blur and JPEG compression with a 0.5 probability for each; (4b) Similar to (4a) but with a 10\% probability. We train our model with the augmented dataset and apply it to the practical dataset $\mathbf{T_{gen}}$. Results in Fig.~\ref{fig:figureaug} show detector performance with various augmentations. DreamBooth \cite{dreambooth2022} maintains high accuracy but decreases with Blur+JPEG(0.5). NightCafe \cite{nightcafe2019} improves with all augmentations. MidjourneyV5 \cite{midjourney2022} fluctuates slightly, decreasing with Blur + JPEG(0.1). StableAI \cite{stabilityai2022} moderately increases in accuracy, while YiJian \cite{yijian} performs the lowest, slightly improving with Blur+JPEG(0.1). These findings highlight how augmentation improves detector performance on real-world datasets.

\subsubsection{\textbf{Spectrum Analysis in RIO Process}} 

\label{Spec-Ana-RIO}
\begin{figure}[htbp]
  \centering
  \includegraphics[width=0.7\columnwidth]{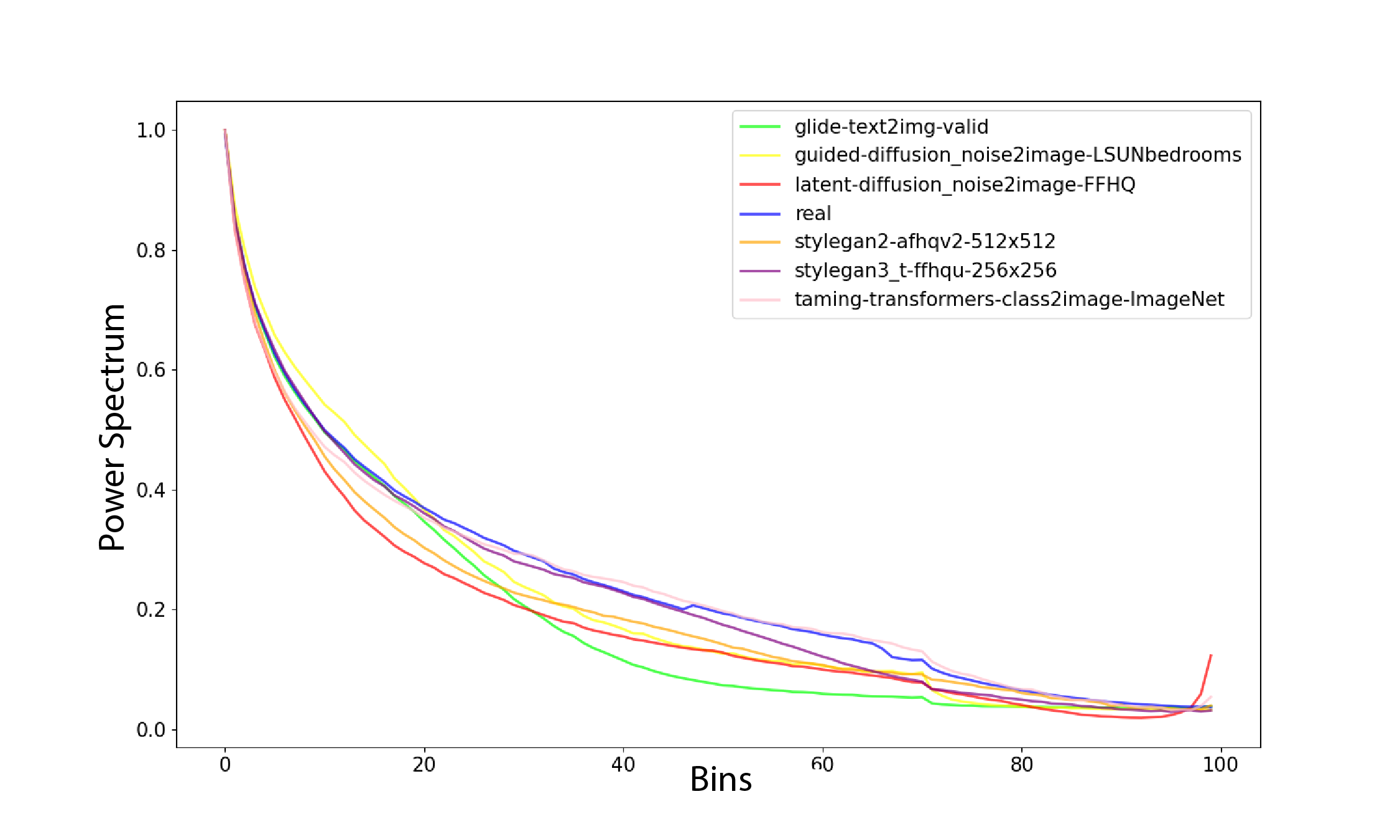}
  \caption{Spectrum Analysis Graph for YCbCr Images after Applying RIO in 10K different images. The graph shows the number of radii on the X-axis and the power spectrum intensity on the Y-axis. This illustrates the power spectrum intensity is different for each generated method in DMs and GANs approaches. The Rigid lines represent the mean for 10K images.}
  \label{fig:figure4.2}
\end{figure}

\vspace{-4pt}
In Figure \ref{fig:figure4.2} distinct images are presented to demonstrate the efficacy of spectrum analysis with our RIO. Specifically, utilizing FFT facilitates clear differentiation between lines generated through various methods. Notably, the line attributed to the ``real'' images underscores heightened fluctuation along the radius in camera-based images. Also, we can observe that the mean for real images fluctuates significantly within a 40 to 80 radius. Furthermore, to enhance clarity, we incorporated diverse generated images and conducted RIO analysis on real and generated images to discern any disparities.


\begin{figure}[htbp]
  \centering
  \includegraphics[width=0.8\columnwidth]{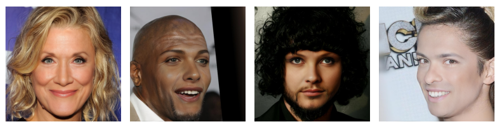}
  \caption{Samples for real-life image's inference on FaceSwap method.}
  \label{fig:figure4.3}
\end{figure}
\subsubsection{\textbf{Inference Time for Unknown Images}}
For effectiveness of real-world scenario, we took more than 3,000 real-world images consisting FaceSwap generation methods, where some samples are presented in Fig. \ref{fig:figure4.3}. Our proposed approach benchmarks at shortest inference time of about 400 milliseconds on average. 

Overall, our ablation study reveals that both SFU and RIO components play a pivotal role in enhancing the model's performance. This enhancement is particularly noticeable due to their integration, which proves highly effective in deeper networks, such as ResNet152. We identify the split and shift mechanism within SFU and the YCbCr conversion in RIO as critical sub-components. Removing these components results in significant drops in accuracy. This comprehensive ablation study shows the importance of these components in improving spatial feature processing and region-specific analysis, affirming our architectural choices.

\section{Deployment and Reproducibility}
Currently, we have deployed our demo system in a live web server
and it has been deployed since Oct. 2023 to help detect real-world deepfakes. Users can upload videos or images, and our approach can provide the results through an analysis tool containing detection results with a CSV file, a frame detection graph, a pie chart, etc. We hope that our tools can be used more widely for people to use to detect AI-generated content. 

\section{Conclusion}
\label{sec:5}
In this work, we introduce \SystemName~to combat new types of AI-generated fake images, where we observed a consistent pattern of frequency distribution among those fake images. Our approach focuses on extracting frequency domain features from the YCbCr color space and introduces Spatial Feature Extraction (SFE) to enhance the frequency features. Our extensive experimental results demonstrate that our approach surpasses the performance of other SOTA methods over various types of AI-generated content, including the latest generative model.
Our work shows a promising avenue for robustly classifying fake images in practical scenarios.

\microtypesetup{activate=false}
\begin{acks}


This work was partly supported by Institute for Information \& communication Technology Planning \& evaluation (IITP) grants funded by the Korean government MSIT:
(RS-2022-II221199, RS-2024-00337703, RS-2022-II220688, RS-2019-II190421, RS-2023-0023 0337, RS-2024-00356293, RS-2022-II221045, and RS-2021-II212068).


\end{acks}
\balance

\bibliographystyle{ACM-Reference-Format}
\bibliography{sample-base}


\begin{thebibliography}{72}


\ifx \showCODEN    \undefined \def \showCODEN     #1{\unskip}     \fi
\ifx \showDOI      \undefined \def \showDOI       #1{#1}\fi
\ifx \showISBNx    \undefined \def \showISBNx     #1{\unskip}     \fi
\ifx \showISBNxiii \undefined \def \showISBNxiii  #1{\unskip}     \fi
\ifx \showISSN     \undefined \def \showISSN      #1{\unskip}     \fi
\ifx \showLCCN     \undefined \def \showLCCN      #1{\unskip}     \fi
\ifx \shownote     \undefined \def \shownote      #1{#1}          \fi
\ifx \showarticletitle \undefined \def \showarticletitle #1{#1}   \fi
\ifx \showURL      \undefined \def \showURL       {\relax}        \fi
\providecommand\bibfield[2]{#2}
\providecommand\bibinfo[2]{#2}
\providecommand\natexlab[1]{#1}
\providecommand\showeprint[2][]{arXiv:#2}

\bibitem[AI(2021)]%
        {danbooru2021}
\bibfield{author}{\bibinfo{person}{Danbooru AI}.} \bibinfo{year}{2021}\natexlab{}.
\newblock \bibinfo{booktitle}{\emph{DanBooru Dataset}}.
\newblock
\urldef\tempurl%
\url{https://gwern.net/danbooru2021}
\showURL{%
\tempurl}


\bibitem[Bang and Woo(2021)]%
        {woo12}
\bibfield{author}{\bibinfo{person}{Young~Oh Bang} {and} \bibinfo{person}{Simon~S Woo}.} \bibinfo{year}{2021}\natexlab{}.
\newblock \showarticletitle{DA-FDFtNet: dual attention fake detection fine-tuning network to detect various AI-generated fake images}.
\newblock \bibinfo{journal}{\emph{arXiv preprint arXiv:2112.12001}} (\bibinfo{year}{2021}).
\newblock


\bibitem[Bergland(1969)]%
        {Bergland}
\bibfield{author}{\bibinfo{person}{G.~D. Bergland}.} \bibinfo{year}{1969}\natexlab{}.
\newblock \showarticletitle{A guided tour of the fast Fourier transform}.
\newblock \bibinfo{journal}{\emph{IEEE Spectrum}} \bibinfo{volume}{6}, \bibinfo{number}{7} (\bibinfo{year}{1969}), \bibinfo{pages}{41--52}.
\newblock
\urldef\tempurl%
\url{https://doi.org/10.1109/MSPEC.1969.5213896}
\showDOI{\tempurl}


\bibitem[Brock et~al\mbox{.}(2019)]%
        {brock2018large}
\bibfield{author}{\bibinfo{person}{Andrew Brock}, \bibinfo{person}{Jeff Donahue}, {and} \bibinfo{person}{Karen Simonyan}.} \bibinfo{year}{2019}\natexlab{}.
\newblock \bibinfo{title}{Large Scale {GAN} Training for High Fidelity Natural Image Synthesis}.
\newblock
\newblock
\urldef\tempurl%
\url{https://openreview.net/forum?id=B1xsqj09Fm}
\showURL{%
\tempurl}


\bibitem[Chai et~al\mbox{.}(2020)]%
        {patchforensics}
\bibfield{author}{\bibinfo{person}{Lucy Chai}, \bibinfo{person}{David Bau}, \bibinfo{person}{Ser-Nam Lim}, {and} \bibinfo{person}{Phillip Isola}.} \bibinfo{year}{2020}\natexlab{}.
\newblock \bibinfo{title}{What makes fake images detectable? Understanding properties that generalize}.
\newblock
\newblock


\bibitem[Chan et~al\mbox{.}(2022)]%
        {Chan2022}
\bibfield{author}{\bibinfo{person}{Eric~R. Chan}, \bibinfo{person}{Connor~Z. Lin}, \bibinfo{person}{Matthew~A. Chan}, \bibinfo{person}{Koki Nagano}, \bibinfo{person}{Boxiao Pan}, \bibinfo{person}{Shalini~De Mello}, \bibinfo{person}{Orazio Gallo}, \bibinfo{person}{Leonidas Guibas}, \bibinfo{person}{Jonathan Tremblay}, \bibinfo{person}{Sameh Khamis}, \bibinfo{person}{Tero Karras}, {and} \bibinfo{person}{Gordon Wetzstein}.} \bibinfo{year}{2022}\natexlab{}.
\newblock \bibinfo{title}{{Efficient Geometry-aware 3D Generative Adversarial Networks}}.
\newblock
\newblock


\bibitem[Chandrasegaran et~al\mbox{.}(2022)]%
        {chandrasegaran2022discovering}
\bibfield{author}{\bibinfo{person}{Keshigeyan Chandrasegaran}, \bibinfo{person}{Ngoc-Trung Tran}, \bibinfo{person}{Alexander Binder}, {and} \bibinfo{person}{Ngai-Man Cheung}.} \bibinfo{year}{2022}\natexlab{}.
\newblock \bibinfo{title}{Discovering Transferable Forensic Features for CNN-generated Images Detection}.
\newblock
\newblock
\showeprint[arxiv]{2208.11342}~[cs.CV]


\bibitem[Cho et~al\mbox{.}(2023)]%
        {woo16}
\bibfield{author}{\bibinfo{person}{Beomsang Cho}, \bibinfo{person}{Binh~M Le}, \bibinfo{person}{Jiwon Kim}, \bibinfo{person}{Simon Woo}, \bibinfo{person}{Shahroz Tariq}, \bibinfo{person}{Alsharif Abuadbba}, {and} \bibinfo{person}{Kristen Moore}.} \bibinfo{year}{2023}\natexlab{}.
\newblock \showarticletitle{Towards understanding of deepfake videos in the wild}. In \bibinfo{booktitle}{\emph{Proceedings of the 32nd ACM International Conference on Information and Knowledge Management}}. \bibinfo{pages}{4530--4537}.
\newblock


\bibitem[Corvi et~al\mbox{.}(2023)]%
        {corvi2023intriguing}
\bibfield{author}{\bibinfo{person}{Riccardo Corvi}, \bibinfo{person}{Davide Cozzolino}, \bibinfo{person}{Giovanni Poggi}, \bibinfo{person}{Koki Nagano}, {and} \bibinfo{person}{Luisa Verdoliva}.} \bibinfo{year}{2023}\natexlab{}.
\newblock \showarticletitle{Intriguing properties of synthetic images: from generative adversarial networks to diffusion models}.
\newblock  (\bibinfo{year}{2023}).
\newblock
\showeprint[arxiv]{2304.06408}~[cs.CV]


\bibitem[Corvi et~al\mbox{.}(2022)]%
        {Corvi2022_on}
\bibfield{author}{\bibinfo{person}{Riccardo Corvi}, \bibinfo{person}{Davide Cozzolino}, \bibinfo{person}{Giada Zingarini}, \bibinfo{person}{Giovanni Poggi}, \bibinfo{person}{Koki Nagano}, {and} \bibinfo{person}{Luisa Verdoliva}.} \bibinfo{year}{2022}\natexlab{}.
\newblock \bibinfo{title}{On the detection of synthetic images generated by diffusion models}.
\newblock
\newblock
\urldef\tempurl%
\url{https://doi.org/10.48550/ARXIV.2211.00680}
\showDOI{\tempurl}


\bibitem[Cozzolino et~al\mbox{.}(2021)]%
        {cozzolino2021universal}
\bibfield{author}{\bibinfo{person}{Davide Cozzolino}, \bibinfo{person}{Diego Gragnaniello}, \bibinfo{person}{Giovanni Poggi}, {and} \bibinfo{person}{Luisa Verdoliva}.} \bibinfo{year}{2021}\natexlab{}.
\newblock \bibinfo{title}{Towards Universal GAN Image Detection}.
\newblock
\newblock
\showeprint[arxiv]{2112.12606}~[cs.CV]


\bibitem[Deng et~al\mbox{.}(2009)]%
        {deng2009imagenet}
\bibfield{author}{\bibinfo{person}{Jia Deng}, \bibinfo{person}{Wei Dong}, \bibinfo{person}{Richard Socher}, \bibinfo{person}{Li-Jia Li}, \bibinfo{person}{Kai Li}, {and} \bibinfo{person}{Li Fei-Fei}.} \bibinfo{year}{2009}\natexlab{}.
\newblock \bibinfo{title}{Imagenet: A large-scale hierarchical image database}.
\newblock , \bibinfo{numpages}{248--255}~pages.
\newblock


\bibitem[Dhariwal and Nichol(2021)]%
        {NEURIPS2021_49ad23d1}
\bibfield{author}{\bibinfo{person}{Prafulla Dhariwal} {and} \bibinfo{person}{Alexander Nichol}.} \bibinfo{year}{2021}\natexlab{}.
\newblock \bibinfo{title}{Diffusion Models Beat GANs on Image Synthesis}.
\newblock , \bibinfo{numpages}{8780--8794}~pages.
\newblock
\urldef\tempurl%
\url{https://proceedings.neurips.cc/paper_files/paper/2021/file/49ad23d1ec9fa4bd8d77d02681df5cfa-Paper.pdf}
\showURL{%
\tempurl}


\bibitem[Esser et~al\mbox{.}(2021)]%
        {esser2021taming}
\bibfield{author}{\bibinfo{person}{Patrick Esser}, \bibinfo{person}{Robin Rombach}, {and} \bibinfo{person}{Bjorn Ommer}.} \bibinfo{year}{2021}\natexlab{}.
\newblock \bibinfo{title}{Taming Transformers for High-Resolution Image Synthesis}.
\newblock
\newblock
\showeprint[arxiv]{2012.09841}~[cs.CV]


\bibitem[Gragnaniello et~al\mbox{.}(2021)]%
        {9428429}
\bibfield{author}{\bibinfo{person}{D. Gragnaniello}, \bibinfo{person}{D. Cozzolino}, \bibinfo{person}{F. Marra}, \bibinfo{person}{G. Poggi}, {and} \bibinfo{person}{L. Verdoliva}.} \bibinfo{year}{2021}\natexlab{}.
\newblock \bibinfo{title}{Are GAN Generated Images Easy to Detect? A Critical Analysis of the State-Of-The-Art}.
\newblock , \bibinfo{numpages}{6}~pages.
\newblock
\urldef\tempurl%
\url{https://doi.org/10.1109/ICME51207.2021.9428429}
\showDOI{\tempurl}


\bibitem[Ho et~al\mbox{.}(2020)]%
        {ho2020denoising}
\bibfield{author}{\bibinfo{person}{Jonathan Ho}, \bibinfo{person}{Ajay Jain}, {and} \bibinfo{person}{Pieter Abbeel}.} \bibinfo{year}{2020}\natexlab{}.
\newblock \bibinfo{title}{Denoising Diffusion Probabilistic Models}.
\newblock
\newblock
\showeprint[arxiv]{2006.11239}~[cs.LG]


\bibitem[Holz(y 12)]%
        {midjourney2022}
\bibfield{author}{\bibinfo{person}{David Holz}.} \bibinfo{year}{2022, July 12}\natexlab{}.
\newblock \bibinfo{booktitle}{\emph{Midjourney: Expanding the Imaginative Powers}}.
\newblock
\urldef\tempurl%
\url{midjourney.com/}
\showURL{%
\tempurl}


\bibitem[Hong et~al\mbox{.}(2024)]%
        {woo13}
\bibfield{author}{\bibinfo{person}{Seunghoo Hong}, \bibinfo{person}{Juhun Lee}, {and} \bibinfo{person}{Simon~S Woo}.} \bibinfo{year}{2024}\natexlab{}.
\newblock \showarticletitle{All but One: Surgical Concept Erasing with Model Preservation in Text-to-Image Diffusion Models}. In \bibinfo{booktitle}{\emph{Proceedings of the AAAI Conference on Artificial Intelligence}}, Vol.~\bibinfo{volume}{38}. \bibinfo{pages}{21143--21151}.
\newblock


\bibitem[Jeon et~al\mbox{.}(2019)]%
        {woo19}
\bibfield{author}{\bibinfo{person}{Hyeonseong Jeon}, \bibinfo{person}{Youngoh Bang}, {and} \bibinfo{person}{Simon~S Woo}.} \bibinfo{year}{2019}\natexlab{}.
\newblock \showarticletitle{Faketalkerdetect: Effective and practical realistic neural talking head detection with a highly unbalanced dataset}. In \bibinfo{booktitle}{\emph{Proceedings of the IEEE/CVF International Conference on Computer Vision Workshops}}. \bibinfo{pages}{0--0}.
\newblock


\bibitem[Jeon et~al\mbox{.}(2020)]%
        {woo27}
\bibfield{author}{\bibinfo{person}{Hyeonseong Jeon}, \bibinfo{person}{Youngoh Bang}, {and} \bibinfo{person}{Simon~S Woo}.} \bibinfo{year}{2020}\natexlab{}.
\newblock \showarticletitle{Fdftnet: Facing off fake images using fake detection fine-tuning network}. In \bibinfo{booktitle}{\emph{IFIP international conference on ICT systems security and privacy protection}}. Springer, \bibinfo{pages}{416--430}.
\newblock


\bibitem[Jeong et~al\mbox{.}(2022)]%
        {frepgan}
\bibfield{author}{\bibinfo{person}{Yonghyun Jeong}, \bibinfo{person}{Doyeon Kim}, \bibinfo{person}{Youngmin Ro}, {and} \bibinfo{person}{Jongwon Choi}.} \bibinfo{year}{2022}\natexlab{}.
\newblock \showarticletitle{FrePGAN: robust deepfake detection using frequency-level perturbations}. In \bibinfo{booktitle}{\emph{Proceedings of the AAAI Conference on Artificial Intelligence}}, Vol.~\bibinfo{volume}{36}. \bibinfo{pages}{1060--1068}.
\newblock


\bibitem[Karras et~al\mbox{.}(2018)]%
        {karras2018progressive}
\bibfield{author}{\bibinfo{person}{Tero Karras}, \bibinfo{person}{Timo Aila}, \bibinfo{person}{Samuli Laine}, {and} \bibinfo{person}{Jaakko Lehtinen}.} \bibinfo{year}{2018}\natexlab{}.
\newblock \bibinfo{title}{Progressive Growing of GANs for Improved Quality, Stability, and Variation}.
\newblock
\newblock
\showeprint[arxiv]{1710.10196}~[cs.NE]


\bibitem[Karras et~al\mbox{.}(2020)]%
        {karras2020analyzing}
\bibfield{author}{\bibinfo{person}{Tero Karras}, \bibinfo{person}{Samuli Laine}, \bibinfo{person}{Miika Aittala}, \bibinfo{person}{Janne Hellsten}, \bibinfo{person}{Jaakko Lehtinen}, {and} \bibinfo{person}{Timo Aila}.} \bibinfo{year}{2020}\natexlab{}.
\newblock \bibinfo{title}{Analyzing and Improving the Image Quality of StyleGAN}.
\newblock
\newblock
\showeprint[arxiv]{1912.04958}~[cs.CV]


\bibitem[Khalid et~al\mbox{.}(2021)]%
        {Khalid2021FakeAVCelebAN}
\bibfield{author}{\bibinfo{person}{Hasam Khalid}, \bibinfo{person}{Shahroz Tariq}, {and} \bibinfo{person}{Simon~S. Woo}.} \bibinfo{year}{2021}\natexlab{}.
\newblock \bibinfo{title}{FakeAVCeleb: A Novel Audio-Video Multimodal Deepfake Dataset}.
\newblock
\newblock
\urldef\tempurl%
\url{https://api.semanticscholar.org/CorpusID:236976127}
\showURL{%
\tempurl}


\bibitem[Khalid and Woo(2020)]%
        {woo20}
\bibfield{author}{\bibinfo{person}{Hasam Khalid} {and} \bibinfo{person}{Simon~S Woo}.} \bibinfo{year}{2020}\natexlab{}.
\newblock \showarticletitle{Oc-fakedect: Classifying deepfakes using one-class variational autoencoder}. In \bibinfo{booktitle}{\emph{Proceedings of the IEEE/CVF conference on computer vision and pattern recognition workshops}}. \bibinfo{pages}{656--657}.
\newblock


\bibitem[Kim et~al\mbox{.}(2019)]%
        {woo22}
\bibfield{author}{\bibinfo{person}{Junyaup Kim}, \bibinfo{person}{Siho Han}, {and} \bibinfo{person}{Simon~S Woo}.} \bibinfo{year}{2019}\natexlab{}.
\newblock \showarticletitle{Classifying genuine face images from disguised face images}. In \bibinfo{booktitle}{\emph{2019 IEEE International Conference on Big Data (Big Data)}}. IEEE, \bibinfo{pages}{6248--6250}.
\newblock


\bibitem[Kim et~al\mbox{.}(2023)]%
        {woo3}
\bibfield{author}{\bibinfo{person}{Jun~Yaup Kim}, \bibinfo{person}{Min~Ha Kim}, {and} \bibinfo{person}{Simon~Sungil Woo}.} \bibinfo{year}{2023}\natexlab{}.
\newblock \bibinfo{title}{Method and Apparatus for Deleting Trained Data of Deep Learning Model}.
\newblock
\newblock
\newblock
\shownote{US Patent App. 17/881,753}.


\bibitem[Kim and Woo(2018)]%
        {woo14}
\bibfield{author}{\bibinfo{person}{Keeyoung Kim} {and} \bibinfo{person}{Simon~S Woo}.} \bibinfo{year}{2018}\natexlab{}.
\newblock \showarticletitle{When George Clooney Is Not George Clooney: Using GenAttack to Deceive Amazon’s and Naver’s Celebrity Recognition APIs}. In \bibinfo{booktitle}{\emph{ICT Systems Security and Privacy Protection: 33rd IFIP TC 11 International Conference, SEC 2018, Held at the 24th IFIP World Computer Congress, WCC 2018, Poznan, Poland, September 18-20, 2018, Proceedings 33}}. Springer, \bibinfo{pages}{355--369}.
\newblock


\bibitem[Kim and Woo(2022)]%
        {woo2}
\bibfield{author}{\bibinfo{person}{Keeyoung Kim} {and} \bibinfo{person}{Simon~S. Woo}.} \bibinfo{year}{2022}\natexlab{}.
\newblock \showarticletitle{Negative Adversarial Example Generation Against Naver's Celebrity Recognition API}. In \bibinfo{booktitle}{\emph{Proceedings of the 1st Workshop on Security Implications of Deepfakes and Cheapfakes}}. \bibinfo{pages}{19--23}.
\newblock


\bibitem[Kim et~al\mbox{.}(2021a)]%
        {woo25}
\bibfield{author}{\bibinfo{person}{Minha Kim}, \bibinfo{person}{Shahroz Tariq}, {and} \bibinfo{person}{Simon~S Woo}.} \bibinfo{year}{2021}\natexlab{a}.
\newblock \showarticletitle{Cored: Generalizing fake media detection with continual representation using distillation}. In \bibinfo{booktitle}{\emph{Proceedings of the 29th ACM International Conference on Multimedia}}. \bibinfo{pages}{337--346}.
\newblock


\bibitem[Kim et~al\mbox{.}(2021b)]%
        {woo31}
\bibfield{author}{\bibinfo{person}{Minha Kim}, \bibinfo{person}{Shahroz Tariq}, {and} \bibinfo{person}{Simon~S Woo}.} \bibinfo{year}{2021}\natexlab{b}.
\newblock \showarticletitle{Fretal: Generalizing deepfake detection using knowledge distillation and representation learning}. In \bibinfo{booktitle}{\emph{Proceedings of the IEEE/CVF conference on computer vision and pattern recognition}}. \bibinfo{pages}{1001--1012}.
\newblock


\bibitem[Le et~al\mbox{.}(2023)]%
        {woo10}
\bibfield{author}{\bibinfo{person}{Binh Le}, \bibinfo{person}{Shahroz Tariq}, \bibinfo{person}{Alsharif Abuadbba}, \bibinfo{person}{Kristen Moore}, {and} \bibinfo{person}{Simon Woo}.} \bibinfo{year}{2023}\natexlab{}.
\newblock \showarticletitle{Why Do Facial Deepfake Detectors Fail?}. In \bibinfo{booktitle}{\emph{Proceedings of the 2nd Workshop on Security Implications of Deepfakes and Cheapfakes}}. \bibinfo{pages}{24--28}.
\newblock


\bibitem[Le et~al\mbox{.}(2024)]%
        {woo11}
\bibfield{author}{\bibinfo{person}{Binh~M Le}, \bibinfo{person}{Jiwon Kim}, \bibinfo{person}{Shahroz Tariq}, \bibinfo{person}{Kristen Moore}, \bibinfo{person}{Alsharif Abuadbba}, {and} \bibinfo{person}{Simon~S Woo}.} \bibinfo{year}{2024}\natexlab{}.
\newblock \showarticletitle{Sok: Facial deepfake detectors}.
\newblock \bibinfo{journal}{\emph{arXiv preprint arXiv:2401.04364}} (\bibinfo{year}{2024}).
\newblock


\bibitem[Le and Woo(2021)]%
        {woo17}
\bibfield{author}{\bibinfo{person}{Binh~M Le} {and} \bibinfo{person}{Simon~S Woo}.} \bibinfo{year}{2021}\natexlab{}.
\newblock \showarticletitle{Exploring the asynchronous of the frequency spectra of gan-generated facial images}.
\newblock \bibinfo{journal}{\emph{arXiv preprint arXiv:2112.08050}} (\bibinfo{year}{2021}).
\newblock


\bibitem[Le and Woo(2023)]%
        {woo21}
\bibfield{author}{\bibinfo{person}{Binh~M Le} {and} \bibinfo{person}{Simon~S Woo}.} \bibinfo{year}{2023}\natexlab{}.
\newblock \showarticletitle{Quality-agnostic deepfake detection with intra-model collaborative learning}. In \bibinfo{booktitle}{\emph{Proceedings of the IEEE/CVF International Conference on Computer Vision}}. \bibinfo{pages}{22378--22389}.
\newblock


\bibitem[Lee et~al\mbox{.}(2024)]%
        {woo34}
\bibfield{author}{\bibinfo{person}{Kangjun Lee}, \bibinfo{person}{Inho Jung}, {and} \bibinfo{person}{Simon~S Woo}.} \bibinfo{year}{2024}\natexlab{}.
\newblock \showarticletitle{iFakeDetector: Real Time Integrated Web-based Deepfake Detection System}.
\newblock  (\bibinfo{year}{2024}).
\newblock


\bibitem[Lee et~al\mbox{.}(2022)]%
        {woo18}
\bibfield{author}{\bibinfo{person}{Sangyup Lee}, \bibinfo{person}{Jaeju An}, {and} \bibinfo{person}{Simon~S Woo}.} \bibinfo{year}{2022}\natexlab{}.
\newblock \showarticletitle{BZNet: unsupervised multi-scale branch zooming network for detecting low-quality deepfake videos}. In \bibinfo{booktitle}{\emph{Proceedings of the ACM Web Conference 2022}}. \bibinfo{pages}{3500--3510}.
\newblock


\bibitem[Lee et~al\mbox{.}(2021a)]%
        {woo23}
\bibfield{author}{\bibinfo{person}{Sangyup Lee}, \bibinfo{person}{Shahroz Tariq}, \bibinfo{person}{Junyaup Kim}, {and} \bibinfo{person}{Simon~S Woo}.} \bibinfo{year}{2021}\natexlab{a}.
\newblock \showarticletitle{Tar: Generalized forensic framework to detect deepfakes using weakly supervised learning}. In \bibinfo{booktitle}{\emph{IFIP International conference on ICT systems security and privacy protection}}. Springer, \bibinfo{pages}{351--366}.
\newblock


\bibitem[Lee et~al\mbox{.}(2021b)]%
        {woo24}
\bibfield{author}{\bibinfo{person}{Sangyup Lee}, \bibinfo{person}{Shahroz Tariq}, \bibinfo{person}{Junyaup Kim}, {and} \bibinfo{person}{Simon~S Woo}.} \bibinfo{year}{2021}\natexlab{b}.
\newblock \showarticletitle{Tar: Generalized forensic framework to detect deepfakes using weakly supervised learning}. In \bibinfo{booktitle}{\emph{IFIP International conference on ICT systems security and privacy protection}}. Springer, \bibinfo{pages}{351--366}.
\newblock


\bibitem[Lee et~al\mbox{.}(2021c)]%
        {woo30}
\bibfield{author}{\bibinfo{person}{Sangyup Lee}, \bibinfo{person}{Shahroz Tariq}, \bibinfo{person}{Youjin Shin}, {and} \bibinfo{person}{Simon~S Woo}.} \bibinfo{year}{2021}\natexlab{c}.
\newblock \showarticletitle{Detecting handcrafted facial image manipulations and GAN-generated facial images using Shallow-FakeFaceNet}.
\newblock \bibinfo{journal}{\emph{Applied soft computing}}  \bibinfo{volume}{105} (\bibinfo{year}{2021}), \bibinfo{pages}{107256}.
\newblock


\bibitem[Li et~al\mbox{.}(2020)]%
        {9156368}
\bibfield{author}{\bibinfo{person}{Yuezun Li}, \bibinfo{person}{Xin Yang}, \bibinfo{person}{Pu Sun}, \bibinfo{person}{Honggang Qi}, {and} \bibinfo{person}{Siwei Lyu}.} \bibinfo{year}{2020}\natexlab{}.
\newblock \bibinfo{title}{Celeb-DF: A Large-Scale Challenging Dataset for DeepFake Forensics}.
\newblock , \bibinfo{numpages}{3204--3213}~pages.
\newblock
\urldef\tempurl%
\url{https://doi.org/10.1109/CVPR42600.2020.00327}
\showDOI{\tempurl}


\bibitem[Lin et~al\mbox{.}(2014)]%
        {lin2014microsoft}
\bibfield{author}{\bibinfo{person}{Tsung-Yi Lin}, \bibinfo{person}{Michael Maire}, \bibinfo{person}{Serge Belongie}, \bibinfo{person}{Lubomir Bourdev}, \bibinfo{person}{Ross Girshick}, \bibinfo{person}{James Hays}, \bibinfo{person}{Pietro Perona}, \bibinfo{person}{Deva Ramanan}, \bibinfo{person}{C~Lawrence Zitnick}, {and} \bibinfo{person}{Piotr Dollar}.} \bibinfo{year}{2014}\natexlab{}.
\newblock \showarticletitle{Microsoft COCO: Common objects in context}.
\newblock \bibinfo{journal}{\emph{European conference on computer vision}} (\bibinfo{year}{2014}), \bibinfo{pages}{740--755}.
\newblock


\bibitem[Ma et~al\mbox{.}(2023)]%
        {ma2023exposing}
\bibfield{author}{\bibinfo{person}{Ruipeng Ma}, \bibinfo{person}{Jinhao Duan}, \bibinfo{person}{Fei Kong}, \bibinfo{person}{Xiaoshuang Shi}, {and} \bibinfo{person}{Kaidi Xu}.} \bibinfo{year}{2023}\natexlab{}.
\newblock \bibinfo{title}{Exposing the Fake: Effective Diffusion-Generated Images Detection}.
\newblock
\newblock
\showeprint[arxiv]{2307.06272}~[cs.CV]


\bibitem[made AI(mber)]%
        {yijian}
\bibfield{author}{\bibinfo{person}{Author:~Tailor made AI}.} \bibinfo{year}{2019, November}\natexlab{}.
\newblock \bibinfo{booktitle}{\emph{Yijian: Chinese AI Painting Creative Cloud}}.
\newblock
\urldef\tempurl%
\url{https://creator.nightcafe.studio/}
\showURL{%
\tempurl}


\bibitem[made AI(2022)]%
        {dreambooth2022}
\bibfield{author}{\bibinfo{person}{Tailor made AI}.} \bibinfo{year}{2022}\natexlab{}.
\newblock \bibinfo{booktitle}{\emph{Dreambooth: Tailor-made AI Image Generation}}.
\newblock
\urldef\tempurl%
\url{https://www.astria.ai/}
\showURL{%
\tempurl}


\bibitem[Mandelli et~al\mbox{.}(2022)]%
        {Mandelli_2022}
\bibfield{author}{\bibinfo{person}{Sara Mandelli}, \bibinfo{person}{Davide Cozzolino}, \bibinfo{person}{Edoardo~D. Cannas}, \bibinfo{person}{Joao~P. Cardenuto}, \bibinfo{person}{Daniel Moreira}, \bibinfo{person}{Paolo Bestagini}, \bibinfo{person}{Walter~J. Scheirer}, \bibinfo{person}{Anderson Rocha}, \bibinfo{person}{Luisa Verdoliva}, \bibinfo{person}{Stefano Tubaro}, {and} \bibinfo{person}{Edward~J. Delp}.} \bibinfo{year}{2022}\natexlab{}.
\newblock \showarticletitle{Forensic Analysis of Synthetically Generated Western Blot Images}.
\newblock \bibinfo{journal}{\emph{{IEEE} Access}}  \bibinfo{volume}{10} (\bibinfo{year}{2022}), \bibinfo{pages}{59919--59932}.
\newblock
\urldef\tempurl%
\url{https://doi.org/10.1109/access.2022.3179116}
\showDOI{\tempurl}


\bibitem[Mostaque(2022)]%
        {stabilityai2022}
\bibfield{author}{\bibinfo{person}{Emad Mostaque}.} \bibinfo{year}{2022}\natexlab{}.
\newblock \bibinfo{booktitle}{\emph{Stability.ai: Stable Diffusion Public Release}}.
\newblock
\urldef\tempurl%
\url{https://stability.ai/blog/stable-diffusion-public-release}
\showURL{%
\tempurl}


\bibitem[Park et~al\mbox{.}(2022)]%
        {woo9}
\bibfield{author}{\bibinfo{person}{Geon-Woo Park}, \bibinfo{person}{Eun-Ju Park}, {and} \bibinfo{person}{Simon~S Woo}.} \bibinfo{year}{2022}\natexlab{}.
\newblock \showarticletitle{Zoom-DF: a dataset for video conferencing deepfake}. In \bibinfo{booktitle}{\emph{Proceedings of the 1st Workshop on Security Implications of Deepfakes and Cheapfakes}}. \bibinfo{pages}{7--11}.
\newblock


\bibitem[Patashnik et~al\mbox{.}(2021)]%
        {patashnik2021stylegannada}
\bibfield{author}{\bibinfo{person}{Or Patashnik}, \bibinfo{person}{Rinon Gal}, \bibinfo{person}{Haggai Maron}, \bibinfo{person}{Gal Chechik}, {and} \bibinfo{person}{Daniel Cohen-Or}.} \bibinfo{year}{2021}\natexlab{}.
\newblock \bibinfo{title}{StyleGAN-NADA: CLIP-Guided Domain Adaptation of Image Generators}.
\newblock
\newblock
\showeprint[arxiv]{2108.00946}~[cs.CV]


\bibitem[Radford et~al\mbox{.}(2021)]%
        {radford2021learning}
\bibfield{author}{\bibinfo{person}{Alec Radford}, \bibinfo{person}{Jong~Wook Kim}, \bibinfo{person}{Chris Hallacy}, \bibinfo{person}{Aditya Ramesh}, \bibinfo{person}{Gabriel Goh}, \bibinfo{person}{Sandhini Agarwal}, \bibinfo{person}{Girish Sastry}, \bibinfo{person}{Amanda Askell}, \bibinfo{person}{Pamela Mishkin}, \bibinfo{person}{Jack Clark}, \bibinfo{person}{Gretchen Krueger}, {and} \bibinfo{person}{Ilya Sutskever}.} \bibinfo{year}{2021}\natexlab{}.
\newblock \bibinfo{title}{Learning Transferable Visual Models From Natural Language Supervision}.
\newblock
\newblock
\showeprint[arxiv]{2103.00020}~[cs.CV]


\bibitem[Ramesh et~al\mbox{.}(2022)]%
        {ramesh2022dalle}
\bibfield{author}{\bibinfo{person}{Aditya Ramesh}, \bibinfo{person}{Mikhail Pavlov}, \bibinfo{person}{Gabriel Goh}, \bibinfo{person}{Scott Gray}, \bibinfo{person}{Alec Radford}, \bibinfo{person}{Mark Chen}, {and} \bibinfo{person}{Ilya Sutskever}.} \bibinfo{year}{2022}\natexlab{}.
\newblock \bibinfo{title}{DALL-E 2}.
\newblock
\newblock
\showeprint[arxiv]{2201.03994}~[cs.CV]


\bibitem[Rombach et~al\mbox{.}(2021)]%
        {rombach2021highresolution}
\bibfield{author}{\bibinfo{person}{Robin Rombach}, \bibinfo{person}{Andreas Blattmann}, \bibinfo{person}{Dominik Lorenz}, \bibinfo{person}{Patrick Esser}, {and} \bibinfo{person}{Björn Ommer}.} \bibinfo{year}{2021}\natexlab{}.
\newblock \bibinfo{title}{High-Resolution Image Synthesis with Latent Diffusion Models}.
\newblock
\newblock
\showeprint[arxiv]{2112.10752}~[cs.CV]


\bibitem[Russell(mber)]%
        {nightcafe2019}
\bibfield{author}{\bibinfo{person}{Angus Russell}.} \bibinfo{year}{2019, November}\natexlab{}.
\newblock \bibinfo{booktitle}{\emph{Nightcafe: Create Amazing Artworks using the Power of Artificial Intelligence}}.
\newblock
\urldef\tempurl%
\url{ttps://creator.nightcafe.studio/}
\showURL{%
\tempurl}


\bibitem[Rössler et~al\mbox{.}(2019)]%
        {9010912}
\bibfield{author}{\bibinfo{person}{Andreas Rössler}, \bibinfo{person}{Davide Cozzolino}, \bibinfo{person}{Luisa Verdoliva}, \bibinfo{person}{Christian Riess}, \bibinfo{person}{Justus Thies}, {and} \bibinfo{person}{Matthias Niessner}.} \bibinfo{year}{2019}\natexlab{}.
\newblock \bibinfo{title}{FaceForensics++: Learning to Detect Manipulated Facial Images}.
\newblock , \bibinfo{numpages}{11}~pages.
\newblock
\urldef\tempurl%
\url{https://doi.org/10.1109/ICCV.2019.00009}
\showDOI{\tempurl}


\bibitem[Son et~al\mbox{.}(2024)]%
        {woo7}
\bibfield{author}{\bibinfo{person}{Geonho Son}, \bibinfo{person}{Juhun Lee}, {and} \bibinfo{person}{Simon~S Woo}.} \bibinfo{year}{2024}\natexlab{}.
\newblock \showarticletitle{Disrupting Diffusion-based Inpainters with Semantic Digression}.
\newblock \bibinfo{journal}{\emph{arXiv preprint arXiv:2407.10277}} (\bibinfo{year}{2024}).
\newblock


\bibitem[Tariq et~al\mbox{.}(2024)]%
        {woo5}
\bibfield{author}{\bibinfo{person}{Razaib Tariq}, \bibinfo{person}{Minji Heo}, \bibinfo{person}{Simon~S Woo}, {and} \bibinfo{person}{Shahroz Tariq}.} \bibinfo{year}{2024}\natexlab{}.
\newblock \showarticletitle{Beyond the Screen: Evaluating Deepfake Detectors under Moire Pattern Effects}. In \bibinfo{booktitle}{\emph{Proceedings of the IEEE/CVF Conference on Computer Vision and Pattern Recognition}}. \bibinfo{pages}{4429--4439}.
\newblock


\bibitem[Tariq et~al\mbox{.}(2022)]%
        {woo26}
\bibfield{author}{\bibinfo{person}{Shahroz Tariq}, \bibinfo{person}{Sowon Jeon}, {and} \bibinfo{person}{Simon~S Woo}.} \bibinfo{year}{2022}\natexlab{}.
\newblock \showarticletitle{Am I a real or fake celebrity? Evaluating face recognition and verification APIs under deepfake impersonation attack}. In \bibinfo{booktitle}{\emph{Proceedings of the ACM Web Conference 2022}}. \bibinfo{pages}{512--523}.
\newblock


\bibitem[Tariq et~al\mbox{.}(2023)]%
        {woo15}
\bibfield{author}{\bibinfo{person}{Shahroz Tariq}, \bibinfo{person}{Sowon Jeon}, {and} \bibinfo{person}{Simon~S Woo}.} \bibinfo{year}{2023}\natexlab{}.
\newblock \showarticletitle{Evaluating trustworthiness and racial bias in face recognition apis using deepfakes}.
\newblock \bibinfo{journal}{\emph{Computer}} \bibinfo{volume}{56}, \bibinfo{number}{5} (\bibinfo{year}{2023}), \bibinfo{pages}{51--61}.
\newblock


\bibitem[Tariq et~al\mbox{.}(2018)]%
        {woo33}
\bibfield{author}{\bibinfo{person}{Shahroz Tariq}, \bibinfo{person}{Sangyup Lee}, \bibinfo{person}{Hoyoung Kim}, \bibinfo{person}{Youjin Shin}, {and} \bibinfo{person}{Simon~S Woo}.} \bibinfo{year}{2018}\natexlab{}.
\newblock \showarticletitle{Detecting both machine and human created fake face images in the wild}. In \bibinfo{booktitle}{\emph{Proceedings of the 2nd international workshop on multimedia privacy and security}}. \bibinfo{pages}{81--87}.
\newblock


\bibitem[Tariq et~al\mbox{.}(2019)]%
        {woo28}
\bibfield{author}{\bibinfo{person}{Shahroz Tariq}, \bibinfo{person}{Sangyup Lee}, \bibinfo{person}{Hoyoung Kim}, \bibinfo{person}{Youjin Shin}, {and} \bibinfo{person}{Simon~S Woo}.} \bibinfo{year}{2019}\natexlab{}.
\newblock \showarticletitle{Gan is a friend or foe? a framework to detect various fake face images}. In \bibinfo{booktitle}{\emph{Proceedings of the 34th ACM/SIGAPP Symposium on Applied Computing}}. \bibinfo{pages}{1296--1303}.
\newblock


\bibitem[Tariq et~al\mbox{.}(2021)]%
        {woo32}
\bibfield{author}{\bibinfo{person}{Shahroz Tariq}, \bibinfo{person}{Sangyup Lee}, {and} \bibinfo{person}{Simon Woo}.} \bibinfo{year}{2021}\natexlab{}.
\newblock \showarticletitle{One detector to rule them all: Towards a general deepfake attack detection framework}. In \bibinfo{booktitle}{\emph{Proceedings of the web conference 2021}}. \bibinfo{pages}{3625--3637}.
\newblock


\bibitem[Wang et~al\mbox{.}(2020)]%
        {wang2020cnngenerated}
\bibfield{author}{\bibinfo{person}{Sheng-Yu Wang}, \bibinfo{person}{Oliver Wang}, \bibinfo{person}{Richard Zhang}, \bibinfo{person}{Andrew Owens}, {and} \bibinfo{person}{Alexei~A. Efros}.} \bibinfo{year}{2020}\natexlab{}.
\newblock \bibinfo{title}{CNN-generated images are surprisingly easy to spot... for now}.
\newblock
\newblock
\showeprint[arxiv]{1912.11035}~[cs.CV]


\bibitem[Wang et~al\mbox{.}(2023)]%
        {wang2023dire}
\bibfield{author}{\bibinfo{person}{Zhendong Wang}, \bibinfo{person}{Jianmin Bao}, \bibinfo{person}{Wengang Zhou}, \bibinfo{person}{Weilun Wang}, \bibinfo{person}{Hezhen Hu}, \bibinfo{person}{Hong Chen}, {and} \bibinfo{person}{Houqiang Li}.} \bibinfo{year}{2023}\natexlab{}.
\newblock \bibinfo{title}{DIRE for Diffusion-Generated Image Detection}.
\newblock
\newblock
\showeprint[arxiv]{2303.09295}~[cs.CV]


\bibitem[Woo et~al\mbox{.}(2022a)]%
        {woo29}
\bibfield{author}{\bibinfo{person}{Simon Woo} {et~al\mbox{.}}} \bibinfo{year}{2022}\natexlab{a}.
\newblock \showarticletitle{Add: Frequency attention and multi-view based knowledge distillation to detect low-quality compressed deepfake images}. In \bibinfo{booktitle}{\emph{Proceedings of the AAAI Conference on Artificial Intelligence}}, Vol.~\bibinfo{volume}{36}. \bibinfo{pages}{122--130}.
\newblock


\bibitem[Woo(2022)]%
        {woo1}
\bibfield{author}{\bibinfo{person}{Simon~S. Woo}.} \bibinfo{year}{2022}\natexlab{}.
\newblock \showarticletitle{Advanced Machine Learning Techniques to Detect Various Types of Deepfakes}. In \bibinfo{booktitle}{\emph{Proceedings of the 1st Workshop on Security Implications of Deepfakes and Cheapfakes}}. \bibinfo{pages}{25--25}.
\newblock


\bibitem[Woo et~al\mbox{.}(2024)]%
        {woo6}
\bibfield{author}{\bibinfo{person}{Simon~Sungil Woo}, \bibinfo{person}{Sang~Yup Lee}, {and} \bibinfo{person}{Jae~Ju AN}.} \bibinfo{year}{2024}\natexlab{}.
\newblock \bibinfo{title}{Low quality deepfake detection device and method of detecting low quality deepfake using the same}.
\newblock
\newblock
\newblock
\shownote{US Patent App. 18/241,317}.


\bibitem[Woo et~al\mbox{.}(2023)]%
        {woo4}
\bibfield{author}{\bibinfo{person}{Simon~Sungil Woo}, \bibinfo{person}{Sang~Yup Lee}, {and} \bibinfo{person}{Shahroz Tariq}.} \bibinfo{year}{2023}\natexlab{}.
\newblock \bibinfo{title}{Apparatus and method for detecting deepfake based on convolutional long short-term memory network}.
\newblock
\newblock
\newblock
\shownote{US Patent App. 18/114,416}.


\bibitem[Woo et~al\mbox{.}(2022b)]%
        {woo8}
\bibfield{author}{\bibinfo{person}{Simon~S Woo}, \bibinfo{person}{Shahroz Tariq}, {and} \bibinfo{person}{Hyoungshick Kim}.} \bibinfo{year}{2022}\natexlab{b}.
\newblock \showarticletitle{WDC'22: 1st Workshop on the Security Implications of Deepfakes and Cheapfakes}. In \bibinfo{booktitle}{\emph{Proceedings of the 2022 ACM on Asia Conference on Computer and Communications Security}}. \bibinfo{pages}{1269--1270}.
\newblock


\bibitem[Wu et~al\mbox{.}(2023)]%
        {wu2023generalizable}
\bibfield{author}{\bibinfo{person}{Haiwei Wu}, \bibinfo{person}{Jiantao Zhou}, {and} \bibinfo{person}{Shile Zhang}.} \bibinfo{year}{2023}\natexlab{}.
\newblock \bibinfo{title}{Generalizable Synthetic Image Detection via Language-guided Contrastive Learning}.
\newblock
\newblock
\showeprint[arxiv]{2305.13800}~[cs.CV]


\bibitem[Zhang et~al\mbox{.}(2023)]%
        {10221905}
\bibfield{author}{\bibinfo{person}{Junbin Zhang}, \bibinfo{person}{Yixiao Wang}, \bibinfo{person}{Hamid~Reza Tohidypour}, {and} \bibinfo{person}{Panos Nasiopoulos}.} \bibinfo{year}{2023}\natexlab{}.
\newblock \bibinfo{title}{Detecting Stable Diffusion Generated Images Using Frequency Artifacts: A Case Study on Disney-Style Art}.
\newblock , \bibinfo{numpages}{1845--1849}~pages.
\newblock
\urldef\tempurl%
\url{https://doi.org/10.1109/ICIP49359.2023.10221905}
\showDOI{\tempurl}


\bibitem[Zhang et~al\mbox{.}(2019)]%
        {zhang2019detecting}
\bibfield{author}{\bibinfo{person}{Xu Zhang}, \bibinfo{person}{Svebor Karaman}, {and} \bibinfo{person}{Shih-Fu Chang}.} \bibinfo{year}{2019}\natexlab{}.
\newblock \bibinfo{title}{Detecting and Simulating Artifacts in GAN Fake Images}.
\newblock
\newblock
\showeprint[arxiv]{1907.06515}~[cs.CV]


\bibitem[Zhu et~al\mbox{.}(2023)]%
        {zhu2023stylegan3}
\bibfield{author}{\bibinfo{person}{Tianlei Zhu}, \bibinfo{person}{Junqi Chen}, \bibinfo{person}{Renzhe Zhu}, {and} \bibinfo{person}{Gaurav Gupta}.} \bibinfo{year}{2023}\natexlab{}.
\newblock \bibinfo{title}{StyleGAN3: Generative Networks for Improving the Equivariance of Translation and Rotation}.
\newblock
\newblock
\showeprint[arxiv]{2307.03898}~[cs.CV]


\end{thebibliography}










\end{document}